\pdfoutput=1

\documentclass[11pt]{article}

\usepackage[final]{ACL2023}

\usepackage{times}
\usepackage{latexsym}
\usepackage{makecell}

\usepackage{fnpct}

\usepackage[T1]{fontenc}

\usepackage[utf8]{inputenc}
\usepackage{amssymb}
\usepackage{amsmath}

\usepackage{enumitem}
\newlist{multianswer}{itemize}{2}
\setlist[multianswer]{label=$\square$, topsep=0pt, itemsep=0pt}
\usepackage{enumitem}
\newlist{mcq}{itemize}{2}
\setlist[mcq]{label=$\bigcirc$, topsep=0pt, itemsep=0pt}
\usepackage{microtype}

\usepackage{inconsolata}

\usepackage{graphicx}
\graphicspath{ {./images/} }

\usepackage{xcolor}
\usepackage{colortbl}

\usepackage{lipsum}
\setlipsum{%
  par-before = \begingroup\color{red},
  par-after = \endgroup
}

\title{What Do Indonesians Really Need from Language Technology?\\A Nationwide Survey} 

\author{Muhammad Dehan Al Kautsar$^1$ \quad Lucky Susanto$^2$ \quad \textbf{Derry Wijaya}$^2$ \quad {\bf Fajri Koto$^1$} \\
$^1$Mohamed bin Zayed University of Artificial Intelligence\\ $^2$Monash University\\
\texttt{\href{mailto:Muhammad.Dehan@mbzuai.ac.ae} {\color{black}{muhammad.dehan@mbzuai.ac.ae}}}
}

\begin{document}
\maketitle
\begin{abstract}

Despite emerging efforts to develop NLP for Indonesia's 700+ local languages, progress remains costly due to the need for direct engagement with native speakers.
However, it is unclear what these language communities truly need from language technology. To address this, we conduct a nationwide survey to assess the actual needs of native Indonesian speakers. Our findings indicate that addressing language barriers, particularly through machine translation and information retrieval, is the most critical priority. Although there is strong enthusiasm for advancements in language technology, concerns around privacy, bias, and the use of public data for AI training highlight the need for greater transparency and clear communication to support broader AI adoption.

\end{abstract}

\section{Introduction}

\begin{figure*}
    \centering
    \small
    \includegraphics[width=\linewidth]{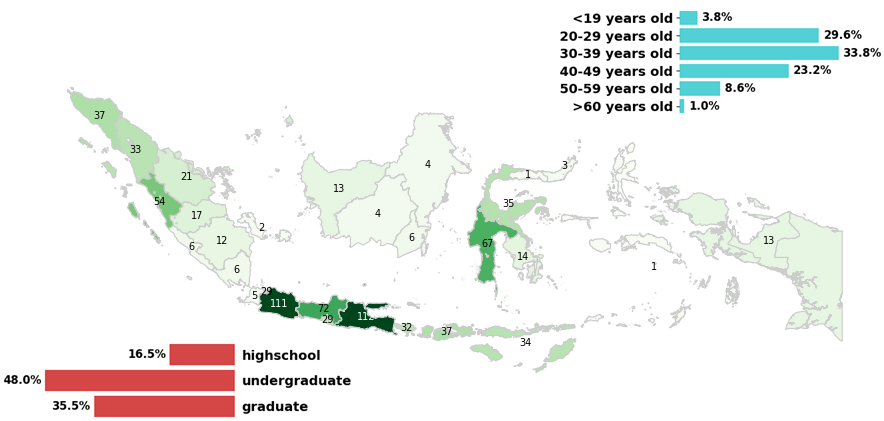}
    \caption{Distribution of respondents by province, along with age and highest education level of local Indonesian language speakers.}
    \label{fig:respondent_distribution}
\end{figure*}



Indonesia, with over 280 million people across 17,508 islands, is home to more than 700 regional languages alongside its national language, Bahasa Indonesia (Indonesian language) \citep{worldbank2024indonesia, ethnologue2024}. While this linguistic diversity offers opportunities for natural language processing (NLP), it also introduces challenges, such as data scarcity and language standardization \citep{novitasari-etal-2020-cross, aji-etal-2022-one}.

To address these challenges, significant efforts have been made in recent years to advance the Indonesian NLP, including multilingual corpora development \citep{cahyawijaya-etal-2023-nusacrowd, seacrowd}, sentiment analysis \citep{winata-etal-2023-nusax}, dialogue \citep{purwarianti-etal-2025-nusadialogue}, and NLU/NLG \citep{koto-etal-2020-indolem,cahyawijaya-etal-2023-nusawrites}. However, the development remains costly and labor-intensive. More importantly, whether these efforts align with actual user needs is still uncertain, leading to a key question: \textit{What do Indonesians truly need from language technologies (LTs)?} Answering this question is essential, as building LTs for Indonesia is particularly complex, partly due to diverse demographics and varying user preferences. Thus, participatory design and engagement with the community are crucial to ensure these technologies serve real-world needs \citep{mager-etal-2023-ethical, kolhatkar-2023-indic, cooper-etal-2024-things}.

To answer these questions and explore the challenges, we conducted a nationwide survey via questionnaire to assess which LTs Indonesians prioritize. We collected demographic data and asked respondents to rate six LTs: Machine Translation (MT), Speech-to-Text (STT), Text-to-Speech (TTS), Grammar Checkers (GC), Information Retrieval (IR), and Digital Assistants (DA). We also examined attitudes toward AI, including concerns about privacy, credibility, and data use~\footnote{We release partial data that contains summarized information and does not reveal any personal information. \href{https://github.com/dehanalkautsar/Indonesian-LT-Survey/}{https://github.com/dehanalkautsar/Indonesian-LT-Survey/}}.
Over two months, we collected 861 responses from speakers of 70 distinct Indonesian languages, representing 35 out of 38 provinces (Figure \ref{fig:respondent_distribution}).

While similar surveys have been conducted in the Global North \cite{blaschke-etal-2024-dialect, lent-etal-2022-creole, soria-etal-2018-dldp}, our findings reveal distinct insights into the needs and concerns of Indonesian language communities. Key findings include: 
\begin{itemize}[noitemsep,topsep=0pt]
    \item LTs bridging language barriers, such as IR and MT, are highly needed. 
    \item Dialects also influence users' interest, demonstrating that the language itself does not solely determine preferences.
    \item 92.6\% of Indonesians are excited about AI technologies, though 36.3\% express concerns.
    \item 86.68\% are aware of potential faults in LTs like DA, but only 46.24\% regularly verify the information provided.
    \item While prior exposure to language technology generally boosts user interest, this trend does not apply uniformly across all demographics, such as Gen-Z and speakers of stable languages, suggesting more complex factors are at play
\end{itemize}

\section{Background and Related Works}

The advancement of NLP is accelerating as the demand for language technologies (LTs) grows \citep{abdalla-etal-2023-elephant}. However, this progress is not evenly distributed worldwide. In Indonesia, NLP development and adoption face significant challenges due to limited resources, linguistic diversity, dialectal and stylistic variations, orthographic inconsistencies, and societal barriers such as unequal access to technology and education across the archipelago \cite{aji-etal-2022-one}. Additionally, as AI technologies evolve, concerns regarding privacy, data collection, and trust add further complexities to development efforts.


\subsection{LTs Surveys Across the World} 
\label{sec:differing-lt-wants}
LT demands vary significantly across regions, reflecting local linguistic, cultural, and technological needs. For instance, a survey of 327 German speakers with dialect found that respondents prioritize dialect-friendly digital assistants over machine translation and spell-checking \citep{blaschke-etal-2024-dialect}. Interviews with Creole experts and 37 people in Creole-speaking communities highlighted speech transcription as a critical unmet need \citep{lent-etal-2022-creol}. Meanwhile, a large-scale survey of over 1,200 speakers of Basque, Breton, Karelian, and Sardinian emphasized the strong desire for language digitalization \citep{soria-etal-2018-dldp}. These examples underscore the diverse and context-dependent nature of LT adoption across the world. 

\citet{millour:hal-02137280} performed a study on European non-standardized language, Alsatian, by designing a series of survey questions and collected responses from over 1,200 participants, most of whom spoke Alsatian and another language, such as French or German. While they successfully identified the state of existing LTs for Alsatians, they did not fully utilize the survey to capture respondents' opinions on available LTs. Similarly, The ELE Project\footnote{\url{https://european-language-equality.eu/deliverables/}},  \citet{mariani:hal-04415222}, and \citet{blasi-etal-2022-systematic} examine the current state and quality of LTs across different languages and demographics, 
but they also lack representation of language speakers' perspectives, leaving their specific LT needs largely unknown.

On the other hand, prior works on ethical considerations have reached the same conclusion when exploring the ethical considerations of building NLP technologies for indigenous languages \cite{bird-2020-decolonising,mager-etal-2023-ethical,kolhatkar-2023-indic,cooper-etal-2024-things}. They recommend that NLP researchers prioritize community engagement rather than solely focusing on de-contextualized artifacts when building NLP technologies. This aligns with our paper's objective of understanding the types of LT needs across the entire Indonesian region—an immense and diverse country with numerous indigenous cultures and languages.

\subsection{Challenges in the Development of LTs in Indonesia}
The development of LTs in Indonesia faces multiple challenges \cite{aji-etal-2022-one}. One primary issue is the lack of resources and the limited awareness of the difficulties faced by underrepresented languages and dialects, e.g., issues with standardization \citep{novitasari-etal-2020-cross}. However, the biggest obstacle remains the availability of sufficient data.

Despite ongoing challenges, researchers and communities have made significant efforts to develop multilingual corpora \citep{cahyawijaya-etal-2023-nusacrowd, seacrowd}, increasing dataset availability and visibility. However, these corpora remain dominated by Indonesian text, with only a small fraction representing local languages. While some datasets emphasize depth (size) \citep{rin-indocia6k, rin-indomakassar9k, rin-inabugi10k} and others prioritize breadth (language coverage) \citep{nllb2022, winata-etal-2023-nusax}, data imbalance persists. In machine translation, only 1.1\% of the 2.3 billion parallel sentences globally involve English-Indonesian pairs, and just 0.06\% cover Javanese-English \citep{gowda-etal-2021-many}. 

Limited data directly affects LT performance, with studies showing significant disparities in LLM capabilities for Indonesian. \citet{koto-etal-2023-indommlu} found that GPT-3.5 struggles with even primary school-level questions in Indonesian and performs worse in regional languages like Sundanese. These challenges in data scarcity and linguistic bias hinder the practical application and commercial viability of LTs in Indonesia. Given these constraints, developing LTs for all Indonesian languages is both costly and complex, highlighting the need to first understand actual user demands before investing in large-scale LT development.

\subsection{Privacy and Bias Issues, alongside Trust in Regards to LTs}
The increasing demand for data to develop language technologies (LTs) has heightened privacy concerns, which have been a longstanding issue even before the emergence of large language models (LLMs). This concern is evident in the implementation of regulatory frameworks such as \citet{GDPR2016} and \citet{CCPA2018}.

Despite regulatory efforts, privacy concerns persist, as research has shown that even anonymized datasets can be vulnerable to re-identification \citep{rocher-deidentify-2019}. This has contributed to growing skepticism toward AI, particularly in Western countries, where only 37\% of Americans believe AI provides more benefits than drawbacks \citep{AIIndex2024Chapter9}. In contrast, attitudes in Indonesia appear more positive, with 78\% of Indonesians viewing AI as beneficial~\citep{AIIndex2024Chapter9}. Differences may influence this optimism in AI exposure, public discourse, and regulatory focus, as discussions on AI ethics and governance are less prominent in these areas compared to Western nations. To better understand public discourse in Indonesia, particularly regarding language technology for local languages, our survey includes questions on perceptions, priorities, and concerns related to AI and LT adoption.

\section{Questionnaire and Data Processing}

\subsection{Questionnaire}

Partially inspired by \citet{blaschke-etal-2024-dialect}, our questionnaire is divided into six sections: introductions, regional language details, opinions on regional languages, LTs-related questions, privacy and credibility of LTs, and respondents' excitement towards AI. The full set of questions is detailed in Appendix \ref{sec:full_questionnaire}, complemented with their answer distributions. The survey is written in \textit{Bahasa Indonesia}, as 94\% of Indonesians understand it, making it easy for respondents to comprehend\footnote {\href{https://translatorswithoutborders.org/wp-content/uploads/2020/04/Indonesian-language-map.pdf}{Indonesian language map}}. Based on follow-up sampling and participant feedback, each respondent required no more than 20 minutes to complete the questionnaire. 

We distributed our questionnaire using \texttt{Google Forms}\footnote{\url{https://docs.google.com/forms}} and shared it through the author's professional networks, reaching language teachers, stakeholders from Indonesian universities, journalists, and local language ambassadors and communities. This approach enabled us to collect responses from across the archipelago, covering 35 out of 38 provinces. Over a window of two months, starting from 06-10-2024 to 05-12-2024, our questionnaire obtained 861 total respondents. Lastly, as a token of appreciation, we randomly award 10 respondents a total of 3,000,000 IDR at the closing time of the questionnaire.

\subsection{Data Processing}
\paragraph{Validating Responses}
To ensure the validity of each response, we require each respondent to share their email address or valid phone number, which is later used for reward selection. Furthermore, our questionnaire also consists of three validation questions that 
require the respondent to either perform a simple addition or select a specific option. These validation questions are randomly embedded throughout the questionnaire, requiring respondents to carefully read each question before responding. These simple validation tasks help detect inattentive responses and prevent bot-generated or random submissions, a method commonly used in large-scale surveys \cite{Muszyski2023AttentionCheck}. After removing responses that do not answer the validation questions correctly, we obtained a total of 811 valid responses, which are used in this work.

\paragraph{Enriching the Responses} 
We enriched the survey responses by considering the respondents' language endangerment level based on \citet{ethnologue2024}. We aggregated their database into a three-tier system: Stable, Threatened, and Moribund, which allows further insights into how language vitality affects the LT needs of the respondents. Further details are available in Appendix \ref{sec:lang_level_aggregation}


\label{sec:response-distribution}
\paragraph{Response Distribution} In total, 811 valid responses were recorded from 35 out of 38 Indonesian provinces, covering 70 of the 700+ languages in Indonesia. With 52.6\% of respondents identifying as women, nearly all participants regularly use technology (computer/laptop/smartphone) in their daily lives, which is crucial given the LT-related questions. 

We aggregated responses based on demographic categories and language endangerment levels. 
Geographically, we collected 574 responses from West Indonesia and 237 from East Indonesia, following the provincial division specified in Appendix \ref{sec:indo_barat_timur}. In terms of generation, 271 respondents belong to Gen-Z, 462 to the millennial generation, and 78 to Gen-X or older.\footnote{Gen Z includes people born in 1997-2010, millennials include those born in 1981-1996, and Gen X or older refers to individuals born before 1980.}

Lastly, based on our aggregation in Appendix \ref{sec:lang_level_aggregation}, respondents were categorized by language endangerment level: 566 as stable language speakers, 196 as threatened language speakers, 17 as moribund language speakers, and 32 as unknowns since their languages do not match any listed in \citet{ethnologue2024}'s local Indonesian languages.

\begin{figure}
    \centering
    \small
    \includegraphics[width=0.98\linewidth]{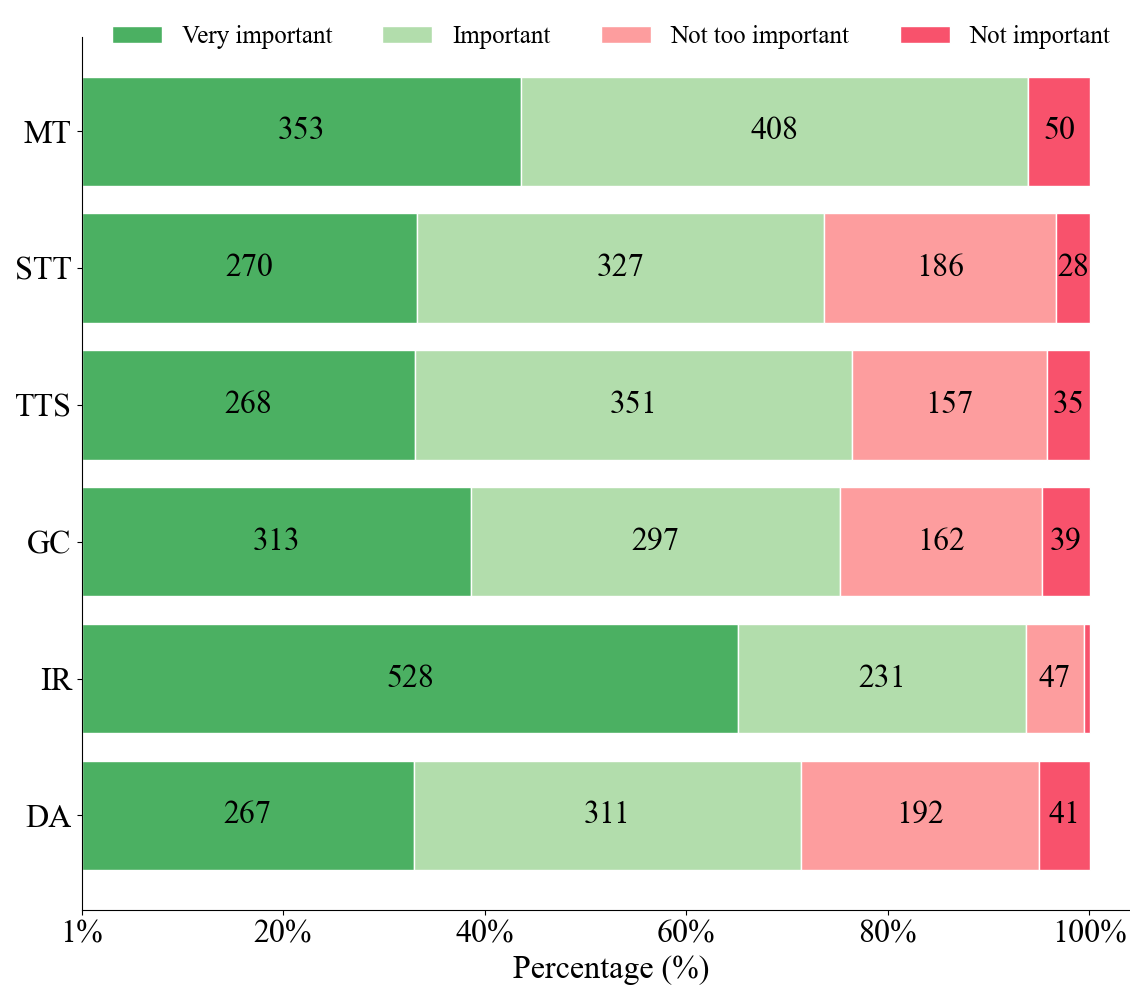}
    \caption{Respondents' views on the importance of various language technologies.}
    \label{fig:main-result}
\end{figure}

\paragraph{Term: Importance Score}
\label{sec:importance-score}
We introduce the term Importance Score (Figure \ref{eq:importance_score}), which helps us quantify how important each LT is based on our respondents' opinions in Section \ref{sec:results}. Respondents rate the importance of each LT based on a 4-level Likert scale: "Very Important", "Important", "Not Very Important", and "Not Important". The Importance Score is a normalization of the weighted value of these responses, where the score of 3 is assigned to "Very Important," decreasing incrementally until "Not Important," which is assigned a score of 0. 

This equation is used to capture respondents’ perceived needs for each LT, as measured in the questionnaire, particularly in questions 23, 26, 29, 33, 36, and 38 (see Appendix~\ref{sec:full_questionnaire}). Using this equation, we are able to conduct analyses based on the respondents’ ordinal categories.

\begin{figure}[htbp]
    \small
    \[
        _\text{Importance Score} = \frac{3N_{\text{VI}} + 2N_{\text{I}} + 1N_{\text{NVI}} + 0N_{\text{NI}}}{3  (N_{\text{VI}} + N_{\text{I}} + N_{\text{NVI}} + N_{\text{NI}})}
    \]
    \caption{How Importance Score \textbf{(IS)} is calculated, values bounded to [0, 1].}
    \label{eq:importance_score}
\end{figure}

\begin{table*}[htbp]
    \centering
    \small
    \begin{tabular}{l|c|c|c|c|c|c|c}
    \hline
    \textbf{Categories} & \textbf{\#} & \textbf{MT} & \textbf{STT} & \textbf{TTS} & \textbf{GC} & \textbf{IR} & \textbf{DA} \\
    \Xhline{4\arrayrulewidth}
    \textbf{full} & \textbf{811} & \textbf{0.771} & \textbf{0.678} & \textbf{0.684} & \textbf{0.696} & \textbf{0.860} & \textbf{0.664} \\
    \Xhline{4\arrayrulewidth}
    aware of bias & 448 & \cellcolor{red!20}-0.70\% & \cellcolor{blue!20}2.06\% & \cellcolor{blue!20}2.25\% & \cellcolor{blue!20}1.88\% & \cellcolor{blue!20}0.88\% & \cellcolor{blue!20}2.08\% \\
    not aware of bias & 363 & \cellcolor{blue!20}0.76\% & \cellcolor{red!20}-2.48\% & \cellcolor{red!50}-2.94\% & \cellcolor{red!20}-2.10\% & \cellcolor{red!20}-1.02\% & \cellcolor{red!20}-2.64\% \\
    \Xhline{4\arrayrulewidth}
    aware of privacy & 467 & \cellcolor{red!20}-1.50\% & \cellcolor{red!20}-0.72\% & \cellcolor{red!20}-0.24\% & \cellcolor{red!20}-0.21\% & \cellcolor{blue!20}0.51\% & \cellcolor{red!20}-0.35\% \\
    not aware of privacy & 344 & \cellcolor{blue!20}1.93\% & \cellcolor{blue!20}1.04\% & \cellcolor{blue!20}0.16\% & \cellcolor{blue!20}0.52\% & \cellcolor{red!20}-0.62\% & \cellcolor{blue!20}0.40\% \\
    \Xhline{4\arrayrulewidth}
    geo: west Indonesia & 574 & \cellcolor{red!20}-1.18\% & \cellcolor{red!50}-3.04\% & \cellcolor{red!50}-3.30\% & \cellcolor{red!50}-2.96\% & \cellcolor{red!20}-1.35\% & \cellcolor{red!50}-4.58\% \\
    geo: east Indonesia & 237 & \cellcolor{blue!50}2.70\% & \cellcolor{blue!50}7.46\% & \cellcolor{blue!50}7.75\% & \cellcolor{blue!50}7.51\% & \cellcolor{blue!50}3.36\% & \cellcolor{blue!70}10.99\% \\
    \Xhline{4\arrayrulewidth}
    edu: highschool & 134 & \cellcolor{red!50}-6.43\% & \cellcolor{red!20}-2.04\% & \cellcolor{red!20}-1.08\% & \cellcolor{red!20}-0.28\% & \cellcolor{blue!20}2.11\% & \cellcolor{blue!20}2.27\% \\
    edu: undergraduate & 389 & \cellcolor{blue!50}2.69\% & \cellcolor{blue!20}1.49\% & \cellcolor{blue!20}0.47\% & \cellcolor{blue!20}0.59\% & \cellcolor{blue!20}0.93\% & \cellcolor{blue!20}1.43\% \\
    edu: graduate & 288 & \cellcolor{red!20}-0.77\% & \cellcolor{red!20}-0.99\% & \cellcolor{red!20}-0.33\% & \cellcolor{red!20}-0.39\% & \cellcolor{red!20}-2.16\% & \cellcolor{red!50}-3.08\% \\
    \Xhline{4\arrayrulewidth}
    lang: stable & 566 & \cellcolor{red!20}-1.08\% & \cellcolor{red!20}-2.28\% & \cellcolor{red!20}-2.28\% & \cellcolor{red!20}-1.76\% & \cellcolor{red!20}-1.94\% & \cellcolor{red!50}-3.32\% \\
    lang: endangered & 196 & \cellcolor{blue!50}4.33\% & \cellcolor{blue!50}7.86\% & \cellcolor{blue!50}5.67\% & \cellcolor{blue!50}6.29\% & \cellcolor{blue!50}4.22\% & \cellcolor{blue!50}8.85\% \\
    lang: moribund & 17 & \cellcolor{red!70}-21.16\% & \cellcolor{red!70}-27.70\% & \cellcolor{red!70}-25.47\% & \cellcolor{red!70}-35.20\% & \cellcolor{blue!20}0.32\% & \cellcolor{red!70}-14.36\% \\
    \Xhline{4\arrayrulewidth}
    familiar with LT & * & \cellcolor{blue!20}0.53\% & \cellcolor{blue!50}5.27\% & \cellcolor{blue!50}7.23\% & \cellcolor{blue!50}4.08\% & \cellcolor{blue!20}0.48\% & \cellcolor{blue!50}6.11\% \\
    not familiar with LT & ** & \cellcolor{red!50}-7.57\% & \cellcolor{red!70}-17.36\% & \cellcolor{red!70}-19.44\% & \cellcolor{red!70}-12.88\% & \cellcolor{red!70}-33.05\% & \cellcolor{red!70}-23.36\% \\
    \Xhline{4\arrayrulewidth}
    gen z & 271 & \cellcolor{red!20}-1.09\% & \cellcolor{red!20}-1.31\% & \cellcolor{blue!20}0.16\% & \cellcolor{blue!20}1.79\% & \cellcolor{blue!20}2.12\% & \cellcolor{blue!50}3.18\% \\
    gen millennial & 462 & \cellcolor{blue!20}0.32\% & \cellcolor{blue!20}1.63\% & \cellcolor{blue!20}0.10\% & \cellcolor{red!20}-1.52\% & \cellcolor{red!20}-0.58\% & \cellcolor{red!20}-0.90\% \\
    gen x boomer & 78 & \cellcolor{blue!20}1.43\% & \cellcolor{red!50}-2.93\% & \cellcolor{red!20}-1.91\% & \cellcolor{blue!50}3.77\% & \cellcolor{red!50}-3.60\% & \cellcolor{red!50}-3.46\% \\
    \Xhline{4\arrayrulewidth}
    \end{tabular}
    \caption{The percentage changes in Language Technologies (LTs) importance scores relative to the overall response across demographic and awareness categories. \textcolor{blue}{Blue} indicates a higher importance score given by respondents compared to the overall response, while \textcolor{red}{red} indicates a lower score. As shown in the table, optimism toward the development of LTs for Indonesian regional languages is primarily driven by respondents from East Indonesia, speakers of endangered languages, and those familiar with LTs. *753, 623, 589, 612, 800, 642 for MT, STT, TTS, GC, IR, DA respectively. **58, 188, 222, 199, 11, 169 for MT, STT, TTS, GC, IR, DA respectively.}
    \label{tab:survey-results}
    \vspace{-1em}
\end{table*}

\paragraph{MT Specific Scoring} We classify respondents' views on the importance of machine translation (\emph{MT}) into three categories: Very Important, Important, and Not Important, to facilitate comparison with other LTs. In the MT importance section, respondents are given six answer choices; five representing different ways MT may be important and one indicating that MT is not important (see Appendix \ref{sec:full_questionnaire} Question 23). We assign '\textit{Very Important}' to respondents who select 3 to 5 options regarding MT’s importance and do not choose Not Important. The `\textit{Important}' category applies to those who select 1 or 2 importance-related options without selecting Not Important. Finally, respondents who choose Not Important are categorized accordingly. The details can be seen in Appendix~\ref{sec:variation-importance-score}.

\section{Results}
\label{sec:results}


\subsection{Which LTs Do Indonesians Need the Most?} 
\label{whichone}
Figure~\ref{fig:main-result} shows that the calculated Importance Score (see Section~\ref{sec:importance-score}) ranks \textbf{IR} highest at 0.860, highlighting its critical role in facilitating information access. In contrast, \textbf{DA} score lowest at 0.664—likely due to limited DA exposure or practical use in regional contexts. Meanwhile, \textbf{MT} leads the mid-range group with a score of 0.771, followed by \textbf{STT}, \textbf{TTS}, and \textbf{GC}. 
Overall, the prominence of IR and MT underscores the importance of bridging linguistic barriers in Indonesia’s linguistically diverse environment \cite{aji-etal-2022-one}.

\subsubsection*{Variations Across Key Categories} 
\label{sec:variation-importance-scores-per-category}
Table~\ref{tab:survey-results} (with additional details in Appendix~\ref{sec:variation-importance-score}) summarizes differences in importance scores across subgroups defined by privacy and bias awareness, LT familiarity, geography, education, language endangerment, and generation. For example, respondents who are aware of privacy issues rate LT needs 0.42\% points lower on average, whereas those who are aware of bias rate them 1.41\% points higher on average. East Indonesian respondents also show a 10.09\% higher preference for DA compared to the overall sample. Generally, they are also more positive with regard to the development of different LTs for their languages compared to West Indonesians. LT familiarity further reinforces support for the development of LTs in their local languages. Similar patterns of positivity also emerge for speakers of endangered languages, though the trend reverses among moribund language speakers. See Sections~\ref{sec:analysis} and~\ref{sec:moribund} for analysis and discussion.

\subsubsection*{MT Direction Needs} 
\begin{figure}
    \centering
    \small
    \includegraphics[width=0.98\linewidth]{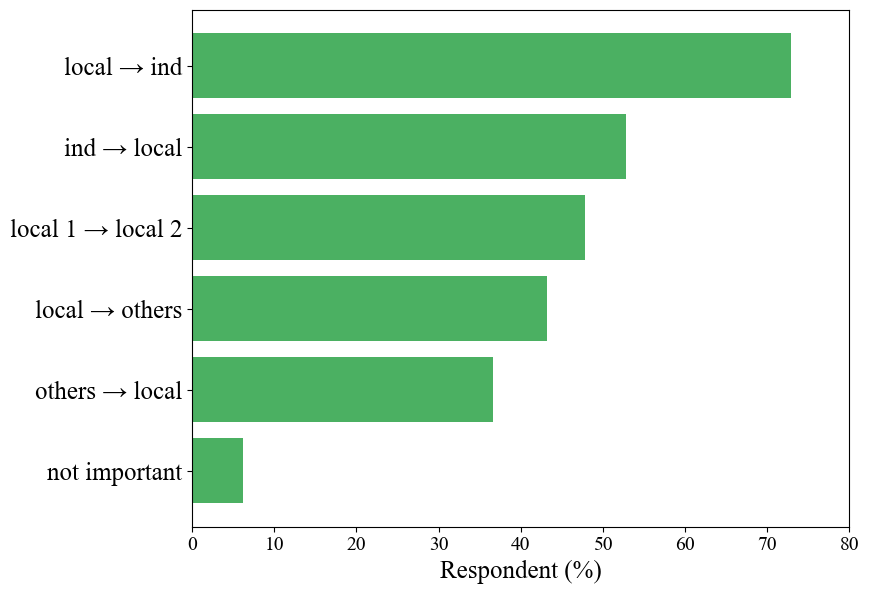}
    \caption{Respondents' views on the importance of machine translation. \textit{local=Indonesian regional language, Ind=Indonesian, others=foreign language.}}
    \label{fig:mt-result}
\end{figure}

As shown in Figure~\ref{fig:mt-result}, the most requested translation direction is from regional languages to Bahasa Indonesia, followed by the reverse. This preference remains consistent across demographics, highlighting Bahasa Indonesia’s role as a unifying medium for inter-regional communication.

\subsection{Dialects Also Influence User Preferences}
\label{sec:dialects}
\begin{figure}
    \centering
    \small
    \includegraphics[width=1\linewidth]{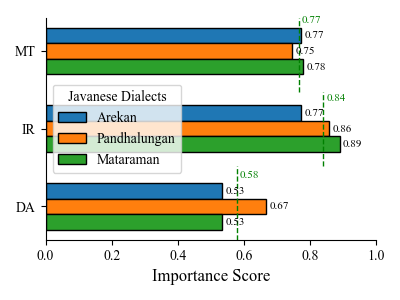}
    \caption{Differences in LT (MT, IR, and DA) preferences across Javanese dialects: Arekan, Pandhalungan, and Mataraman. The dashed line indicates the average among the groups.}
    \label{fig:jav-dialect-want-hbar}
\end{figure}

Our findings reveal that differences in user preferences are not solely based on demographic categories but also arise within the same language due to dialectal variations. Figure \ref{fig:jav-dialect-want-hbar} highlights the differences in LT preferences among speakers of three Javanese dialects: Arekan, Pandhalungan, and Mataraman. The result shows that Javanese speakers of the Pandhalungan dialect express a stronger preference for DA compared to other dialects but show less interest in MT. Additionally, speakers of the Mataraman dialect prioritize information retrieval IR. A detailed analysis of dialectal differences in other languages is provided in Appendix \ref{sec:dialect-appendix}, highlighting that LT preferences can vary even among speakers of the same language.

\subsection{How AI Issues Affect Indonesians' Excitement About AI Technology}
Our survey reveals that 92.6\% of respondents expressed excitement about AI technologies, reflecting a generally optimistic attitude toward technological advancements. However, only 36.3\% of respondents expressed concerns about the development of AI technology, which is lower than the 66\% reported by \citet{AIIndex2024Chapter9}. 
Notably, concerns about AI are closely linked to respondents' awareness of specific issues such as privacy and bias.

\subsubsection*{Privacy Issues}
We directly asked respondents about their awareness of privacy issues and their opinions on the matter in the questionnaire (see Appendix \ref{sec:full_questionnaire}, questions 42 and 43). 
Awareness of privacy issues appears to strongly influence concerns about AI. Among the 197 respondents who believe there are no privacy issues in current AI technology, only \textbf{53 (26.9\%)} expressed concerns about AI. In contrast, among the 363 respondents who believe privacy issues exist, \textbf{163 (44.9\%)} reported concerns. Lastly, among the 251 respondents who were unaware of privacy issues, \textbf{79 (31.4\%)} expressed concerns.
These findings suggest that individuals who recognize privacy issues are more likely to be apprehensive about AI technologies, highlighting privacy as a key factor shaping public perception.

\subsubsection*{Bias Issues}
A similar trend is observed regarding bias in AI technology.
As with privacy issues, we asked respondents about their awareness of bias in LTs, explicitly providing examples of bias in the questionnaire (see Appendix \ref{sec:full_questionnaire}, question 48). 
Among the 157 respondents who were unaware of bias issues, only \textbf{41 (26.1\%)} expressed concerns about AI. In contrast, among the 654 respondents who were aware of bias issues, \textbf{254 (38.8\%)} expressed concerns.
These results suggest that awareness of bias increases recognition of potential risks in AI, though its impact on concern appears to be lower compared to privacy issues.

\subsection{Indonesians' Awareness of Fact-checking Necessities}
\begin{figure}
    \centering
    \includegraphics[width=\linewidth]{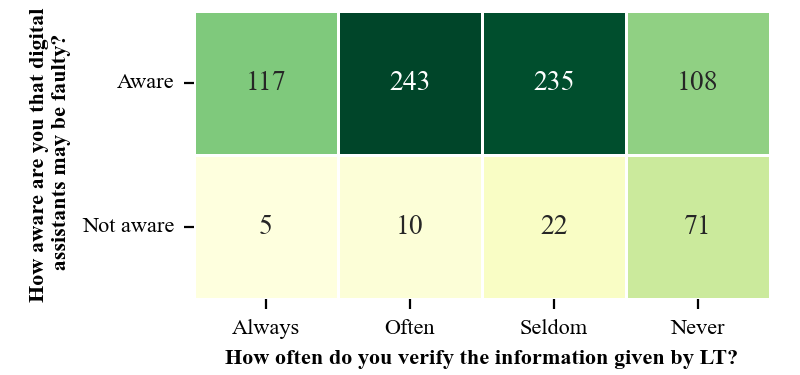}
    \caption{Heatmap of how awareness of LT's hallucination affects respondents' trust.}
    \label{fig:aware-fact-affects-trust}
\end{figure}

Figure \ref{fig:aware-fact-affects-trust} illustrates the trend of how awareness of LLM's hallucination influences respondents' tendency to fact-check information. Based on our survey, \textbf{86.68\%} of respondents are aware that LTs, such as digital assistants, may be flawed and provide incorrect or non-factual information. However, despite this high level of awareness, only \textbf{46.24\%} of respondents regularly verify the information provided by LTs, highlighting how our respondents perceive and respond to the unreliability of the LT-generated information.

Furthermore, when considering only respondents who do not regularly verify information from LTs, we find that \textbf{19.50\% }of them have asked LTs about health-related issues, in contrast to \textbf{48.27\%} of respondents who have inquired about health problems \textit{and} also regularly fact-check the information they receive. This suggests that individuals who do not routinely verify information may be less likely to use LTs for fact-sensitive inquiries. Additionally, concerns about data privacy make individuals more cautious about sharing personal information, such as health conditions, due to fears that current AI systems may not adequately protect their data, as detailed in Appendix \ref{sec:privacy-use-rate}.


\subsection{Does Prior Exposure to LT Influence LT Needs?}
\label{sec:analysis}

\begin{figure}
    \centering
    \small
    \includegraphics[width=0.98\linewidth]{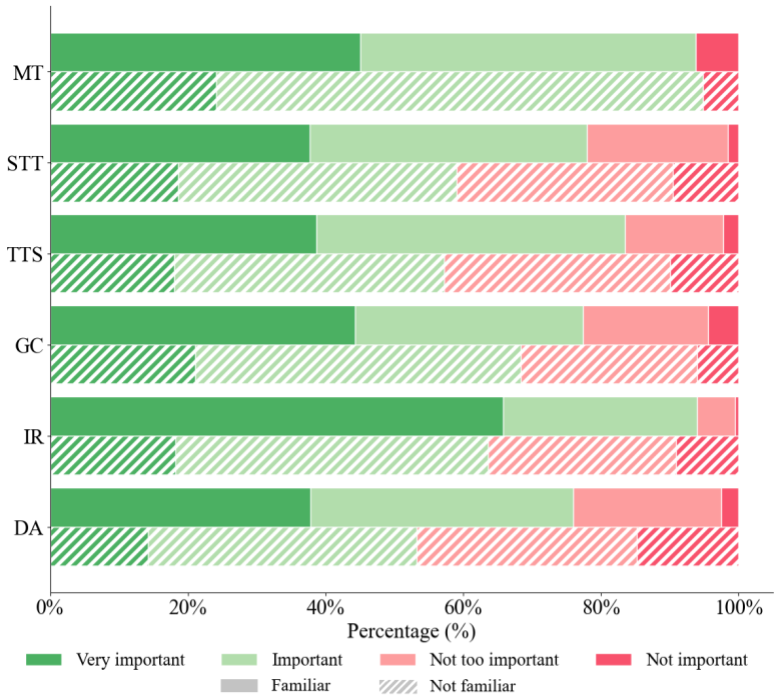}
    \caption{Respondents' views on the importance of various LTs split by familiarity.}
    \label{fig:4-main-result-familiarity}
\end{figure}

Respondents with little to no exposure to a specific LT are more likely to perceive it as unimportant. This trend holds across all LTs except for machine translation, which remains highly valued regardless of familiarity (Figure \ref{fig:4-main-result-familiarity}). 


Furthermore, Appendix \ref{sec:familiarity-familiar} examines how respondents' familiarity with a specific LT influences the importance they assign to the development of the LT in their local language (and the correlation between their familiarity and these perceived importance). According to the Pearson correlation analysis (Figure \ref{fig:importance-score-gen-langlvl-westeast}, Appendix \ref{sec:familiarity-familiar}), certain groups—such as \textit{Gen-X/Boomers} show a strong positive correlation between their familiarity with IR and the importance they place on IR. Similarly, the \textit{Moribund language speakers} show a strong positive correlation between their familiarity and perceived importance of TTS and DA. In addition, familiarity with and perceived importance of TTS and DA consistently exhibit strong positive correlations across different demographic categories. This suggests a shared behavioral pattern and a relationship between respondents' familiarity with these technologies and their perceived importance.


However, despite younger generations, such as Gen-Z, and speakers of stable languages having greater familiarity with language technologies (refer to Figure \ref{fig:familiar-generation-familiar}, more details in Appendix~\ref{sec:dialect-appendix}), the importance scores they assign to the LT are not always the highest within the LT category. 
This suggests that while familiarity with LTs influences perceptions of their importance, it does not always dictate their prioritization. These findings raise intriguing questions about other underlying factors driving these perceptions that remain unexplored in this study.

\begin{figure*}
    \centering
    \includegraphics[width=0.98\linewidth]{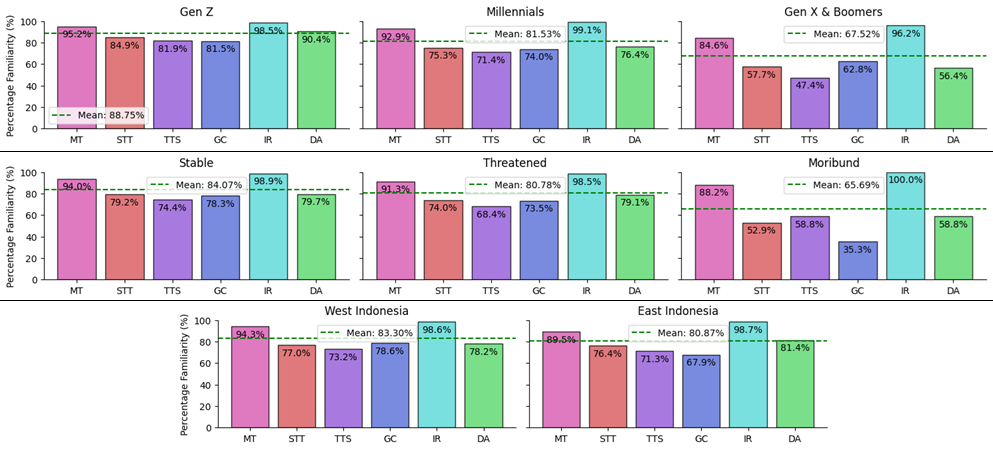}
    \caption{Familiarity with LTs by multiple categories. The top row categorizes data by generation (Gen Z, Millennials, Gen X \& Boomers), the middle row by language endangerment level, and the bottom row by Indonesian region (West and East Indonesia).}
    \label{fig:familiar-generation-familiar}
\end{figure*}


\section{Discussion}
\paragraph{Limited Regional Data as a Barrier to LT Development}
Appendix \ref{sec:important-vs-resourcewiki} demonstrates that while respondents consider language technologies (LTs) to be highly important, the availability of data poses a barrier to their development, especially for underrepresented regional languages. For instance, respondents from the Bugis community, consisting of 4 million speakers\footnote{\url{https://www.ethnologue.com/language/bug}}, strongly encourage the development of language technologies (LTs). However,  existing training data for the Bugis language is limited to less than 10 MB\footnote{We calculate the size based on the Bugis Wikipedia page.}, which severely hinders technological advancements. Similarly, we observe that endangered language speakers are on average more excited for the development of LTs in their languages (Table \ref{tab:survey-results}). Unfortunately, there are also languages with limited data. As shown in Figure \ref{fig:importance-vs-resourcewiki}, Appendix \ref{sec:important-vs-resourcewiki}, some of the languages with the most excitement, such as Bugis, Toba, and Aceh, are among the languages with the lowest existing resources.

Moreover, as shown in Appendix \ref{sec:lt-curr-state}, the current state of LT development for real-world applications reveals a disparity. While higher-resource languages like Javanese are increasingly integrated into LTs, many low-resource languages with substantial speaker populations remain unsupported. 
This underscores a critical challenge in advancing LTs for Indonesia’s regional languages—without adequate data, progress in natural language processing (NLP) applications remains constrained.

\paragraph{Indonesian LT Needs Are Driven by Language Barriers}
As anticipated, language technology (LT) preferences vary across geopolitical regions. Compared to other countries (see Section \ref{sec:differing-lt-wants}), Indonesians' LT priorities appear to be strongly influenced by language barriers, with Information Retrieval (IR) and Machine Translation (MT) being the most highly valued. This aligns with Indonesia's vast linguistic diversity, which, while culturally enriching, also poses information access and communication challenges. In this context, LTs have the potential to serve as unifying tools, transforming linguistic diversity from a barrier into a national strength, a sentiment shared by previous works such as \citet{aji-etal-2022-one}. A key finding is that Indonesians strongly desire search engines to support regional languages. This result can be substantial to the future development of Indonesian LTs.

\paragraph{Are There Concerns in the Use of Public Data?}
Our survey revealed that 11\% of respondents expressed opposition to the use of public data, either text or audio, for the development of language technologies (LTs) supporting regional languages. Further analysis showed that this percentage is not influenced by factors such as respondents' awareness of privacy or bias issues, their excitement about or concerns for AI technologies, or the endangerment status of their language. These findings suggest that concerns about public data usage may stem from factors beyond the scope of the variables considered in our study. Further investigation is needed to uncover the underlying reasons for these reservations among Indonesians, which could include cultural sensitivities, trust in institutions, data colonialism concerns \cite{couldry2019data}, or specific experiences with data misuse or digital labor issues 
\cite{le2023problem}. 


\paragraph{Why Moribund Language Speakers Aren't As Excited About LTs} 
\label{sec:moribund}
Table \ref{tab:survey-results} reveals that unlike endangered language speakers who show the most enthusiasm for LTs, speakers of Moribund languages show less enthusiasm for developing LTs in their local languages. We hypothesize that this attitude stems from their limited understanding of the language's current state and the perception that it no longer serves as a practical means of communication. To explore this further, we interviewed a government official responsible for revitalization of endangered and threatened languages, who cited the Beilel language as an example of a language community that has declined offers from the Indonesian government for revitalization efforts. With only five sibling pairs who barely understand the language, they no longer see its practical utility and primarily use more accessible languages for communication, such as Kabola (\textit{klz})\footnote{Kabola is classified as endangered by \citet{ethnologue2024}.}\footnote{For more details, see \href{https://www.rri.co.id/atambua/daerah/866260/tidak-ada-penutur-bahasa-beilel-terancam-punah}{RRI News}}. This suggests that while LTs can support language revitalization efforts, their impact may be limited to languages that are still classified as endangered. Once a language reaches a Moribund state, securing community support for revitalization becomes more challenging. This underscores the urgent need for dedicated research and the development of relevant LTs before a language reaches this critical stage.

\section{Conclusion}


In this study, we surveyed 35 out of 38 provinces in Indonesia, gathering over 800 responses to assess public attitudes toward Language Technologies (LTs). Our findings underscore a strong national priority for LTs that facilitate access to information and inter-regional communication, particularly through information retrieval (IR) and machine translation (MT). These technologies are essential for overcoming linguistic barriers and ensuring digital inclusivity. 

Additionally, we observe a high level of enthusiasm for AI technologies among respondents, though this is coupled with concerns regarding privacy, bias, and the use of public data for training LTs. Given that prior familiarity with LTs correlates with a higher perception of their importance, increasing public exposure and education on LTs could help address these concerns, fostering greater trust and widespread adoption.

Our analysis and interview also highlight the urgent need to develop LTs and linguistic resources while communities are still engaged. Waiting too long risks missing the window of opportunity, as languages that decline into a Moribund state often lose community support for revitalization efforts. Developing LTs for regional languages before they reach this critical stage is vital to ensuring their continued functionality in society and preserving Indonesia’s rich linguistic diversity. Dedicated research is necessary to prevent these languages from becoming irretrievably lost, making the development of LTs not just beneficial but imperative.

\section*{Limitations}
Our results represent a sample of the Indonesian population, with the majority of respondents being stable language speakers, millennials, residents of West Indonesia, undergraduates, and already familiar with certain LTs. The use of an online platform also limits representation for those without access to such technology. While this means our findings may not capture every possible perspective, the responses are far from uniform. The diverse range of inputs allows for a detailed analysis as presented in Section \ref{sec:results}. Additionally, to ensure transparency, we provide a breakdown of respondent distribution in Section \ref{sec:response-distribution}, with each demographic category further analyzed in Section \ref{sec:variation-importance-scores-per-category}.

We encountered challenges in finding moribund language speakers for our survey, managing to collect only 17 out of 811 valid responses. Due to the sparse distribution and tiny amount of moribund language speakers across Indonesia, reaching them proved difficult. To address this, we maximized respondent collection efforts, hoping to include as many moribund language speakers as possible.

In the questionnaire, even though we adopted attention-check questions \cite{Muszyski2023AttentionCheck}, there was still a possibility that some respondents attempted to fill out the survey multiple times to increase their chances of winning the prize. To further mitigate this, we implemented an additional safeguard by identifying duplicate phone numbers or emails. If duplicates were found, only one response was retained, and the respondent was deemed ineligible for the prize.

Furthermore, in the MT importance question, instead of asking respondents what type of MT they consider important, as done in question 23 of Appendix \ref{sec:full_questionnaire}, we could have structured the question similarly to those for other LTs. However, we designed it this way to gain a clearer understanding of which aspects of MT are most relevant to their daily lives.

\section*{Ethical Consideration}
We only collected data from respondents who consented to its use for further analysis. 
At the beginning of the survey (see Appendix \ref{sec:full_questionnaire}), we provided clear information about the survey's purpose, explicitly stating that it is an academic study with no commercial intent and assured respondents that their personal data would be kept confidential and used solely for research purposes, by ensuring that the data related with the fine-grained information and repository remain private under all circumstances. However, the collective insights are published in the author's repository, protected under the \href{https://creativecommons.org/licenses/by-nc-sa/4.0/}{CC BY-NC-SA 4.0 License}.

However, the participants were not fully anonymized, as we requested contact information to implement a raffle system for rewards/prizes—a common practice in Indonesia to show appreciation. That said, providing contact details was not mandatory; participants could skip that section and still complete the survey. Additionally, apart from the demographic information used for deeper analysis, we did not collect other sensitive data (e.g., name, specific location) to maintain the privacy of the respondents while still conducting comprehensive research.

\section*{Acknowledgements}
This research was funded by MBZUAI Research Grants, Lembaga Pengelola Dana Pendidikan (LPDP), the Ministry of Education, Culture, Research and Technology of the Republic of Indonesia through the Indonesia-US Research Collaboration in Open Digital Technology program, and Monash University’s Action Lab. Their support for this research is deeply appreciated, and we acknowledge their vital role in the successful completion of this work. The findings and conclusions presented in this publication are those of the authors and do not necessarily reflect the views of the sponsors.

\bibliography{anthology,custom}
\bibliographystyle{acl_natbib}

\appendix

\section{Full Questionnaire}
\label{sec:full_questionnaire}
In this section, we present the full questionnaire in its original Indonesian wording, followed by the English translation. The original text is highlighted in \textbf{black}, while the translation is in \textcolor{gray}{\textit{grey-italic}}, and additional details in \textcolor{blue}{blue}. Furthermore, Attention-check questions \cite{Muszyski2023AttentionCheck} and our method to validate the responses are marked in \textcolor{red}{red}.

\rule{0.2\textwidth}{0.4pt}
\vspace{5pt}

\noindent \textbf{Survei Teknologi Bahasa untuk Bahasa-Bahasa Daerah di Indonesia} \newline
\textcolor{gray}{\textit{Language Technology (LT) Survey for Indonesian Local Languages}}

\rule{0.2\textwidth}{0.4pt}
\vspace{5pt}

\noindent Survei ini dilakukan untuk memahami pemahaman masyarakat terkait teknologi bahasa untuk bahasa-bahasa daerah di Indonesia. Survei ini merupakan penelitian akademik dan tidak bersifat komersil. \newline 
Teknologi bahasa berbasis kecerdasan buatan (AI) seperti Google Translate, Google Assistant, dan Siri sudah sering kita gunakan dalam kehidupan sehari-hari. Survei ini bertujuan untuk mengetahui pendapat Anda tentang penggunaan teknologi bahasa untuk bahasa daerah Anda. Survei ini ditujukan bagi Anda yang memiliki kemampuan berbahasa daerah. Kerahasiaan data responden akan dijaga dengan baik dan hanya  akan digunakan untuk keperluan survei ini. \newline 
Total hadiah yang disediakan adalah Rp 3.000.000,-. Di akhir survei (pada tanggal 8 Desember 2024), kami akan memilih 10 pemenang secara acak yang akan mendapatkan masing-masing Rp 300.000,- \newline
\textcolor{gray}{\textit{This survey was conducted to understand the public's understanding of LT for regional languages in Indonesia. This survey is an academic research and is not commercial in nature. \newline Artificial intelligence (AI)-based LT such as Google Translate, Google Assistant, and Siri are often used in our daily lives. This survey aims to find out your opinion on the use of LT for your regional language. This survey is intended for those of you who have regional language skills. The confidentiality of respondent data will be well maintained and will only be used for the purposes of this survey. \newline The total prize provided is IDR 3,000,000. At the end of the survey (on December 8, 2024), we will randomly select 10 winners who will each receive IDR 300,000.}}

\rule{0.2\textwidth}{0.4pt}
\vspace{5pt}

\noindent 1. Apakah Anda bisa menggunakan bahasa daerah? (Pilih semua yang sesuai) \newline
\textcolor{gray}{\textit{1. Can you use any regional language? (Select all that apply)}}
\begin{multianswer}
    \item Saya bisa berbicara menggunakan bahasa daerah \textcolor{blue}{747 (86.8\%)}  \newline
    \textcolor{gray}{\textit{I can speak using regional language}}
    \item Saya bisa menulis dengan bahasa daerah \textcolor{blue}{533 (61.9\%)} \newline
    \textcolor{gray}{\textit{I can write using regional language}} 
    \item Saya bisa membaca dan memahami teks dengan bahasa daerah \textcolor{blue}{652 (75.7\%)} \newline
    \textcolor{gray}{\textit{I can read and understand text in regional language}}
    \item Saya tidak bisa sama sekali \textcolor{blue}{30 (3.5\%)} \newline
    \textcolor{gray}{\textit{I cannot}}
\end{multianswer}

\rule{0.2\textwidth}{0.4pt}
\vspace{5pt}

\noindent \textbf{Perkenalan diri} \newline
\textcolor{gray}{\textit{Introduction}}
\vspace{10pt}

\noindent 2. Tuliskan bahasa daerah yang Anda kuasai! \newline
\textcolor{gray}{\textit{2. Write any regional languages that you are adept with!}} \newline
\textcolor{blue}{861 write-in answers}
\vspace{10pt}

\noindent 3. Tuliskan dialek bahasa daerah Anda (jika ada)! \newline Dialek adalah variasi bahasa yang digunakan oleh sekelompok penutur dengan ciri-ciri tertentu, seperti letak geografis daerah dan ciri-ciri yang relatif sama. \newline Contoh: (1) dialek Toba, (2) dialek Mandailing, (3) dialek Simalungun, (4) dialek Pakpak (Dairi), (5) dialek Karo. \newline
\textcolor{gray}{\textit{3. Write down your regional language dialect (if any)! \newline Dialect is a variation of a language used by a group of speakers with certain characteristics, such as the geographical location of the area and relatively similar characteristics. \newline Examples: (1) Toba dialect, (2) Mandailing dialect, (3) Simalungun dialect, (4) Pakpak (Dairi) dialect, (5) Karo dialect.}} \newline
\textcolor{blue}{838 write-in answers. 23 people answer `-' or `tidak ada' (\textit{no dialect})}
\vspace{10pt}

\noindent 4. Seberapa fasih Anda menggunakan bahasa daerah? \newline
\textcolor{gray}{\textit{4. How fluent are you in your regional language?}}
\begin{mcq}
    \item Sangat fasih \textcolor{blue}{289 (33.6\%)} \newline
    \textcolor{gray}{\textit{Very fluent}}
    \item Fasih \textcolor{blue}{449 (52.1\%)} \newline
    \textcolor{gray}{\textit{Fluent}}
    \item Tidak fasih \textcolor{blue}{110 (12.8\%)} \newline
    \textcolor{gray}{\textit{Not fluent}}
    \item Sangat tidak fasih \textcolor{blue}{13 (1.5\%)} \newline
    \textcolor{gray}{\textit{Very not fluent}}
\end{mcq}
\vspace{15pt}

\noindent 5. Seberapa sering Anda menggunakan bahasa daerah? \newline
\textcolor{gray}{\textit{5. How often do you use your regional language?}}
\begin{mcq}
    \item Setiap hari \textcolor{blue}{534 (62\%)} \newline
    \textcolor{gray}{\textit{Everyday}}
    \item Beberapa kali dalam seminggu \textcolor{blue}{205 (23.8\%)} \newline
    \textcolor{gray}{\textit{A few times a week}}
    \item Sekali dalam seminggu \textcolor{blue}{26 (3\%)} \newline
    \textcolor{gray}{\textit{Once a week}}
    \item Sekali dalam sebulan \textcolor{blue}{16 (1.9\%)} \newline
    \textcolor{gray}{\textit{Once a month}}
    \item Sangat jarang \textcolor{blue}{80 (9.3\%)} \newline
    \textcolor{gray}{\textit{Very rarely}}
\end{mcq}
\vspace{15pt}

\noindent 6. Dari provinsi mana Anda berasal? \newline
\textcolor{gray}{\textit{6. Which province are you from?}} \newline
\textcolor{blue}{multiple choice question with 38 provinces as the radio options. 861 answers}
\vspace{10pt}

\noindent 7. Apa suku Anda? (Jika tidak memiliki suku Anda dapat menuliskan "Indonesia") \newline
\textcolor{gray}{\textit{7. What is your tribe? (you can write "Indonesia" if not any)}} \newline
\textcolor{blue}{861 write-in answers. 46 people answer `Indonesia'}
\vspace{10pt}

\noindent 8. Apa jenis kelamin Anda? \newline
\textcolor{gray}{\textit{8. What is your gender?}}
\begin{mcq}
    \item Perempuan \textcolor{blue}{453 (52.6\%)} \newline
    \textcolor{gray}{\textit{Female}}
    \item Laki-laki \textcolor{blue}{408 (47.4\%)} \newline
    \textcolor{gray}{\textit{Male}}
\end{mcq}
\vspace{15pt}

\noindent 9. Apa pendidikan terakhir Anda? \newline
\textcolor{gray}{\textit{9. What is your last level of education?}}
\begin{mcq}
    \item Tidak bersekolah \textcolor{blue}{1 (0.1\%)} \newline
    \textcolor{gray}{\textit{Did not attend school}}
    \item SD \textcolor{blue}{0 (0\%)} \newline
    \textcolor{gray}{\textit{Elementary school}}
    \item SMP \textcolor{blue}{0 (0\%)} \newline
    \textcolor{gray}{\textit{Junior high school}}
    \item SMA \textcolor{blue}{144 (16.7\%)} \newline
    \textcolor{gray}{\textit{Senior high school}}
    \item S1 \textcolor{blue}{412 (47.9\%)} \newline
    \textcolor{gray}{\textit{Undergraduate}}
    \item S2 \textcolor{blue}{257 (29.8\%)} \newline
    \textcolor{gray}{\textit{Graduate}}
    \item S3 \textcolor{blue}{47 (5.5\%)} \newline
    \textcolor{gray}{\textit{Doctoral}}
\end{mcq}
\vspace{15pt}

\noindent 10. Berapa usia Anda? \newline
\textcolor{gray}{\textit{10. How old are you?}}
\begin{mcq}
    \item <19 tahun \textcolor{blue}{34 (3.9\%)} \newline
    \textcolor{gray}{\textit{Less than 19 years old}}
    \item 20-29 tahun \textcolor{blue}{251 (29.2\%)} \newline
    \textcolor{gray}{\textit{20-29 years old}}
    \item 30-39 tahun \textcolor{blue}{290 (33.7\%)} \newline
    \textcolor{gray}{\textit{30-39 years old}}
    \item 40-49 tahun \textcolor{blue}{195 (22.6\%)} \newline
    \textcolor{gray}{\textit{40-49 years old}}
    \item 50-59 tahun \textcolor{blue}{80 (9.3\%)} \newline
    \textcolor{gray}{\textit{50-59 years old}}
    \item >60 tahun \textcolor{blue}{11 (1.3\%)} \newline
    \textcolor{gray}{\textit{>60 years old}}
\end{mcq}
\vspace{15pt}

\noindent 11. Apa pekerjaan Anda? \newline
\textcolor{gray}{\textit{11. What is your occupation?}} \newline
\textcolor{blue}{861 write-in answers.}
\vspace{10pt}

\noindent 12. Pada situasi apa saja Anda menggunakan bahasa daerah secara aktif (menulis, berbicara) maupun secara pasif (membaca, mendengar)? \newline
\textcolor{gray}{\textit{12. In what type of situations do you use your regional language, either actively (writing, speaking) or passively (reading, listening)}}
\begin{multianswer}
    \item Pesan singkat seperti SMS, WhatsApp, dan sejenisnya \textcolor{blue}{564 (65.5\%)}  \newline
    \textcolor{gray}{\textit{Text message e.g. SMS, WhatsApp, etc.}}
    \item Postingan sosial media \textcolor{blue}{207 (24\%)} \newline
    \textcolor{gray}{\textit{Social media posts}} 
    \item Kolom komentar sosial media \textcolor{blue}{203 (23.6\%)} \newline
    \textcolor{gray}{\textit{Social media comments}}
    \item Percakapan sehari-hari \textcolor{blue}{726 (84.3\%)} \newline
    \textcolor{gray}{\textit{Daily conversations}}
    \item Karya sastra/seni \textcolor{blue}{80 (9.3\%)} \newline
    \textcolor{gray}{\textit{Literary/artistic work}}
    \item Catatan pribadi \textcolor{blue}{135 (15.7\%)} \newline
    \textcolor{gray}{\textit{Personal notes}}
    \item Lainnya \textcolor{blue}{150 write-in answers} \newline
    \textcolor{gray}{\textit{Other}}
\end{multianswer}
\vspace{15pt}

\noindent 13. Isikan nomor WhatsApp atau email Anda. (untuk menghubungi Anda jika Anda memenangkan undian) \newline
\textcolor{gray}{\textit{13. Fill in your WhatsApp number or email. (for contact purposes if you won the raffle)}} \newline
\textcolor{blue}{861 write-in answers. (2 responses are duplicated, \textcolor{red}{so we omit one response and keep the other})}
\vspace{10pt}

\noindent 14. \textcolor{red}{Berapa seratus ditambah seratus?} \newline
\textcolor{gray}{\textit{14. How much is one hundred plus one hundred?}}
\begin{mcq}
    \item Seratus* \textcolor{blue}{8 (0.9\%)} \newline
    \textcolor{gray}{\textit{One hundred}}
    \item Dua ratus \textcolor{blue}{847 (98.4\%)} \newline
    \textcolor{gray}{\textit{Two hundred}}
    \item Tiga ratus* \textcolor{blue}{2 (0.2\%)} \newline
    \textcolor{gray}{\textit{Three hundred}}
    \item Empat ratus* \textcolor{blue}{4 (0.5\%)} \newline
    \textcolor{gray}{\textit{Four hundred}}
\end{mcq}
\textit{note: \textcolor{red}{*we omit these responses from analysis}}

\rule{0.2\textwidth}{0.4pt}
\vspace{5pt}

\noindent \textbf{Pertanyaan Berkaitan dengan Bahasa Daerah} \newline
\textcolor{gray}{\textit{Questions Related to Regional Languages}} \newline
\noindent Isi beberapa pertanyaan berikut dengan mengondisikan Anda dan bahasa daerah Anda pada beberapa pernyataan di bawah ini. \newline
\textcolor{gray}{\textit{Fill these questions by conditioning you and your local language in some statements below.}}
\vspace{10pt}

\noindent 15. Bahasa daerah saya memiliki variasi tingkat kesopanan, seperti perbedaan kata saat berbicara dengan sebaya dan orang yang lebih tua. \newline
\textcolor{gray}{\textit{15. My regional language has some politeness variations level, like the different use of words when talking with people of the same age and older ones.}}
\begin{mcq}
    \item Ya \textcolor{blue}{799 (92.8\%)} \newline
    \textcolor{gray}{\textit{Yes}}
    \item Tidak \textcolor{blue}{44 (5.1\%)} \newline
    \textcolor{gray}{\textit{No}}
    \item Tidak tahu \textcolor{blue}{18 (2.1\%)} \newline
    \textcolor{gray}{\textit{Do not know}}
\end{mcq}
\vspace{15pt}

\noindent 16. Saya sering menjumpai bahasa daerah saya digunakan dalam percakapan langsung. \newline
\textcolor{gray}{\textit{16. I often encounter my regional language used in verbal conversations.}}
\begin{mcq}
    \item Sangat setuju \textcolor{blue}{487 (56.6\%)} \newline
    \textcolor{gray}{\textit{Highly agree}}
    \item Setuju \textcolor{blue}{343 (39.8\%)} \newline
    \textcolor{gray}{\textit{Agree}}
    \item Tidak setuju \textcolor{blue}{28 (3.3\%)} \newline
    \textcolor{gray}{\textit{Disagree}}
    \item Sangat tidak setuju \textcolor{blue}{3 (0.3\%)} \newline
    \textcolor{gray}{\textit{Highly disagree}}
\end{mcq}
\vspace{15pt}

\noindent 17. Saya sering menjumpai bahasa daerah saya dalam bentuk tulisan. \newline
\textcolor{gray}{\textit{17. I often encounter my regional language used in written form.}}
\begin{mcq}
    \item Sangat setuju \textcolor{blue}{210 (24.4\%)} \newline
    \textcolor{gray}{\textit{Highly agree}}
    \item Setuju \textcolor{blue}{417 (48.4\%)} \newline
    \textcolor{gray}{\textit{Agree}}
    \item Tidak setuju \textcolor{blue}{212 (24.6\%)} \newline
    \textcolor{gray}{\textit{Disagree}}
    \item Sangat tidak setuju \textcolor{blue}{22 (2.6\%)} \newline
    \textcolor{gray}{\textit{Highly disagree}}
\end{mcq}
\vspace{15pt}

\rule{0.2\textwidth}{0.4pt}
\vspace{5pt}

\noindent \textbf{Sikap terhadap Bahasa Daerah} \newline
\textcolor{gray}{\textit{Attitude Towards Local Languages}} \newline
\noindent Isi beberapa pertanyaan berikut dengan mengondisikan Anda pada beberapa pernyataan di bawah ini. \newline
\textcolor{gray}{\textit{Fill these questions by conditioning you in some statements below.}}
\vspace{10pt}

\noindent 18. Saya ingin bahasa daerah tetap lestari dan digunakan oleh banyak orang. \newline
\textcolor{gray}{\textit{18. I want regional languages to remain sustainable and used by many people.}}
\begin{mcq}
    \item Sangat setuju \textcolor{blue}{675 (78.4\%)} \newline
    \textcolor{gray}{\textit{Highly agree}}
    \item Setuju \textcolor{blue}{179 (20.8\%)} \newline
    \textcolor{gray}{\textit{Agree}}
    \item Tidak setuju \textcolor{blue}{5 (0.6\%)} \newline
    \textcolor{gray}{\textit{Disagree}}
    \item Sangat tidak setuju \textcolor{blue}{2 (0.2\%)} \newline
    \textcolor{gray}{\textit{Highly disagree}}
\end{mcq}
\vspace{15pt}

\noindent 19. Saya ingin belajar bahasa daerah lain di Indonesia. \newline
\textcolor{gray}{\textit{19. I want to learn other regional languages in Indonesia.}}
\begin{mcq}
    \item Sangat setuju \textcolor{blue}{402 (46.7\%)} \newline
    \textcolor{gray}{\textit{Highly agree}}
    \item Setuju \textcolor{blue}{420 (48.8\%)} \newline
    \textcolor{gray}{\textit{Agree}}
    \item Tidak setuju \textcolor{blue}{38 (4.4\%)} \newline
    \textcolor{gray}{\textit{Disagree}}
    \item Sangat tidak setuju \textcolor{blue}{1 (0.1\%)} \newline
    \textcolor{gray}{\textit{Highly disagree}}
\end{mcq}
\vspace{15pt}

\noindent 20. Saya sering menjumpai orang-orang dengan bahasa daerah, akan tetapi saya tidak bisa memahami bahasa mereka. \newline
\textcolor{gray}{\textit{20. I often meet people with regional languages, but I can't understand their language.}}
\begin{mcq}
    \item Sangat setuju \textcolor{blue}{243 (28.2\%)} \newline
    \textcolor{gray}{\textit{Highly agree}}
    \item Setuju \textcolor{blue}{512 (59.5\%)} \newline
    \textcolor{gray}{\textit{Agree}}
    \item Tidak setuju \textcolor{blue}{102 (11.8\%)} \newline
    \textcolor{gray}{\textit{Disagree}}
    \item Sangat tidak setuju \textcolor{blue}{4 (0.5\%)} \newline
    \textcolor{gray}{\textit{Highly disagree}}
\end{mcq}

\rule{0.2\textwidth}{0.4pt}
\vspace{5pt}

\noindent \textbf{Pertanyaan Berkaitan dengan Teknologi Bahasa} \newline
\textcolor{gray}{\textit{Questions Related to Language Technology}} \newline
\vspace{5pt}

\noindent 21. Apakah aksara bahasa daerah Anda sudah didukung oleh teknologi seperti smartphone atau komputer? \newline
\textcolor{gray}{\textit{21. Is your regional language script supported by technology such as smartphones or computers?}}
\begin{mcq}
    \item Ya \textcolor{blue}{291 (33.8\%)} \newline
    \textcolor{gray}{\textit{Yes}}
    \item Tidak \textcolor{blue}{365 (42.4\%)} \newline
    \textcolor{gray}{\textit{No}}
    \item Tidak tahu \textcolor{blue}{205 (23.8\%)} \newline
    \textcolor{gray}{\textit{Do not know}}
\end{mcq}
\vspace{5pt}

\noindent \textbf{Mesin Penerjemah} \newline
\textcolor{gray}{\textit{Machine Translation}}
\vspace{10pt}

\noindent 22. Apakah Anda pernah menggunakan mesin penerjemah, seperti Google Translate? \newline
\textcolor{gray}{\textit{22. Have you ever used a translation machine, such as Google Translate?}}
\begin{mcq}
    \item Ya \textcolor{blue}{792 (92.0\%)} \newline
    \textcolor{gray}{\textit{Yes}}
    \item Tidak \textcolor{blue}{69 (8.0\%)} \newline
    \textcolor{gray}{\textit{No}}
\end{mcq}
\vspace{15pt}

\noindent 23. Seberapa pentingkah mesin penerjemah bahasa daerah untuk kebutuhan Anda? \newline
\textcolor{gray}{\textit{23. How important is a regional language translation machine for your needs?}}
\begin{multianswer}
    \item Penting untuk menerjemahkan bahasa daerah ke bahasa Indonesia. \textcolor{blue}{622 (72.2\%)} \newline
    \textcolor{gray}{\textit{It is important to translate regional languages into Indonesian.}}
    \item Penting untuk menerjemahkan bahasa Indonesia ke bahasa daerah. \textcolor{blue}{454 (52.7\%)} \newline
    \textcolor{gray}{\textit{It is important to translate Indonesian into regional languages.}}
    \item Penting untuk menerjemahkan antar bahasa daerah. \textcolor{blue}{410 (47.6\%)} \newline
    \textcolor{gray}{\textit{It is important to translate between regional languages.}}
    \item Penting untuk menerjemahkan bahasa daerah ke bahasa asing. \textcolor{blue}{374 (43.4\%)} \newline
    \textcolor{gray}{\textit{It is important to translate regional languages into foreign languages.}}
    \item Penting untuk menerjemahkan bahasa asing ke bahasa daerah. \textcolor{blue}{33 (3.8\%)} \newline
    \textcolor{gray}{\textit{It is important to translate foreign languages into regional languages.}}
    \item Tidak penting \textcolor{blue}{52 (6.0\%)} \newline
    \textcolor{gray}{\textit{Not important}}
\end{multianswer}
\vspace{15pt}

\noindent 24. Dimana Anda ingin melihat atau menggunakan mesin penerjemah untuk bahasa daerah Anda? \newline
\textcolor{gray}{\textit{24. Where would you like to see or use a translation machine for your regional language?}}
\begin{multianswer}
    \item Aplikasi ponsel \textcolor{blue}{668 (77.6\%)} \newline
    \textcolor{gray}{\textit{Mobile apps}}
    \item Platform sosial media \textcolor{blue}{267 (31.0\%)} \newline
    \textcolor{gray}{\textit{Social media platforms}}
    \item Situs web \textcolor{blue}{454 (52.7\%)} \newline
    \textcolor{gray}{\textit{Websites}}
    \item Dokumen digital (PDF, word) \textcolor{blue}{151 (17.5\%)} \newline
    \textcolor{gray}{\textit{Digital documents (PDF, word)}}
    \item Platform pembelajaran online \textcolor{blue}{192 (22.3\%)} \newline
    \textcolor{gray}{\textit{Online learning platforms}}
    \item Sistem di tempat kerja \textcolor{blue}{114 (13.2\%)} \newline
    \textcolor{gray}{\textit{Workplace systems}}
    \item Saat bepergian atau di tempat umum \textcolor{blue}{282 (32.8\%)} \newline
    \textcolor{gray}{\textit{While traveling or in public}}
    \item Tidak tertarik \textcolor{blue}{25 (2.9\%)} \newline
    \textcolor{gray}{\textit{Not interested}}
\end{multianswer}
\vspace{5pt}

\noindent \textbf{Speech-to-text} \newline
\textcolor{gray}{\textit{Speech-to-text}}
\vspace{10pt}

\noindent 25. Speech-to-text adalah sistem yang bisa merubah suara menjadi teks. Apakah Anda pernah menggunakan aplikasi ini? \newline
\textcolor{gray}{\textit{25. Speech-to-text is a system that converts speech into text. Have you ever used an application like this?}}
\begin{mcq}
    \item Ya \textcolor{blue}{655 (76.1\%)} \newline
    \textcolor{gray}{\textit{Yes}}
    \item Tidak \textcolor{blue}{206 (23.9\%)} \newline
    \textcolor{gray}{\textit{No}}
\end{mcq}
\vspace{15pt}

\noindent 26. Seberapa pentingkah speech-to-text bahasa daerah untuk kebutuhan Anda? \newline
\textcolor{gray}{\textit{26. How important is regional language text-to-speech for your needs?}}
\begin{mcq}
    \item Sangat penting \textcolor{blue}{285 (33.1\%)} \newline
    \textcolor{gray}{\textit{Very important}}
    \item Penting \textcolor{blue}{349 (40.5\%)} \newline
    \textcolor{gray}{\textit{Important}}
    \item Tidak terlalu penting \textcolor{blue}{197 (22.9\%)} \newline
    \textcolor{gray}{\textit{Not very important}}
    \item Tidak penting \textcolor{blue}{30 (3.5\%)} \newline
    \textcolor{gray}{\textit{Not important}}
\end{mcq}
\vspace{15pt}

\noindent 27. Dimana Anda ingin melihat atau menggunakan speech-to-text untuk bahasa daerah Anda? \newline
\textcolor{gray}{\textit{27. Where would you like to see or use speech-to-text for your regional language?}}
\begin{multianswer}
    \item Aplikasi ponsel \textcolor{blue}{684 (79.4\%)} \newline
    \textcolor{gray}{\textit{Mobile apps}}
    \item Platform sosial media \textcolor{blue}{246 (28.6\%)} \newline
    \textcolor{gray}{\textit{Social media platforms}}
    \item Situs web \textcolor{blue}{358 (41.6\%)} \newline
    \textcolor{gray}{\textit{Websites}}
    \item Dokumen digital (PDF, word) \textcolor{blue}{131 (15.2\%)} \newline
    \textcolor{gray}{\textit{Digital documents (PDF, word)}}
    \item Platform pembelajaran online \textcolor{blue}{183 (21.3\%)} \newline
    \textcolor{gray}{\textit{Online learning platforms}}
    \item Sistem di tempat kerja \textcolor{blue}{119 (13.8\%)} \newline
    \textcolor{gray}{\textit{Workplace systems}}
    \item Saat bepergian atau di tempat umum \textcolor{blue}{249 (28.9\%)} \newline
    \textcolor{gray}{\textit{While traveling or in public}}
    \item Tidak tertarik \textcolor{blue}{58 (6.7\%)} \newline
    \textcolor{gray}{\textit{Not interested}}
\end{multianswer}
\vspace{5pt}

\noindent \textbf{Text-to-speech} \newline
\textcolor{gray}{\textit{Text-to-speech}}
\vspace{10pt}

\noindent 28. Text-to-speech adalah sistem yang mengubah teks menjadi suara. Apakah Anda pernah menggunakan aplikasi seperti ini? \newline
\textcolor{gray}{\textit{28. Text-to-speech is a system that converts text into speech. Have you ever used an application like this?}}
\begin{mcq}
    \item Ya \textcolor{blue}{620 (72.0\%)} \newline
    \textcolor{gray}{\textit{Yes}}
    \item Tidak \textcolor{blue}{241 (28.0\%)} \newline
    \textcolor{gray}{\textit{No}}
\end{mcq}
\vspace{15pt}

\noindent 29. Seberapa pentingkah text-to-speech bahasa daerah untuk kebutuhan Anda? \newline
\textcolor{gray}{\textit{29. How important is regional language text-to-speech for your needs?}}
\begin{mcq}
    \item Sangat penting \textcolor{blue}{283 (32.9\%)} \newline
    \textcolor{gray}{\textit{Very important}}
    \item Penting \textcolor{blue}{373 (43.3\%)} \newline
    \textcolor{gray}{\textit{Important}}
    \item Tidak terlalu penting \textcolor{blue}{168 (19.5\%)} \newline
    \textcolor{gray}{\textit{Not very important}}
    \item Tidak penting \textcolor{blue}{37 (4.3\%)} \newline
    \textcolor{gray}{\textit{Not important}}
\end{mcq}
\vspace{15pt}

\noindent 30. Dimana Anda ingin melihat atau menggunakan text-to-speech untuk bahasa daerah Anda? \newline
\textcolor{gray}{\textit{30. Where would you like to see or use text-to-speech for your regional language?}}
\begin{multianswer}
    \item Aplikasi ponsel \textcolor{blue}{691 (80.3\%)} \newline
    \textcolor{gray}{\textit{Mobile apps}}
    \item Platform sosial media \textcolor{blue}{283 (32.9\%)} \newline
    \textcolor{gray}{\textit{Social media platforms}}
    \item Situs web \textcolor{blue}{392 (45.5\%)} \newline
    \textcolor{gray}{\textit{Websites}}
    \item Dokumen digital (PDF, word) \textcolor{blue}{145 (16.8\%)} \newline
    \textcolor{gray}{\textit{Digital documents (PDF, word)}}
    \item Platform pembelajaran online \textcolor{blue}{172 (20.0\%)} \newline
    \textcolor{gray}{\textit{Online learning platforms}}
    \item Sistem di tempat kerja \textcolor{blue}{123 (14.3\%)} \newline
    \textcolor{gray}{\textit{Workplace systems}}
    \item Saat bepergian atau di tempat umum \textcolor{blue}{250 (29.0\%)} \newline
    \textcolor{gray}{\textit{While traveling or in public}}
    \item Tidak tertarik \textcolor{blue}{50 (5.8\%)} \newline
    \textcolor{gray}{\textit{Not interested}}
\end{multianswer}
\vspace{15pt}

\noindent 31. \textcolor{red}{Pilih jawaban yang merupakan nama warna} \newline
\textcolor{gray}{\textit{31. Choose the answer that is the name of a color}}
\begin{mcq}
    \item Baju* \textcolor{blue}{11 (1.3\%)} \newline
    \textcolor{gray}{\textit{Clothes}}
    \item Perahu* \textcolor{blue}{0 (0.0\%)} \newline
    \textcolor{gray}{\textit{Boat}}
    \item Merah \textcolor{blue}{846 (98.3\%)} \newline
    \textcolor{gray}{\textit{Red}}
    \item Kursi* \textcolor{blue}{1 (0.1\%)} \newline
    \textcolor{gray}{\textit{Chair}}
    \item Pena* \textcolor{blue}{3 (0.3\%)} \newline
    \textcolor{gray}{\textit{Pen}}
\end{mcq}
\textit{note: \textcolor{red}{*we omit these responses from analysis}}
\vspace{15pt}

\noindent \textbf{Grammar Checkers} \newline
\textcolor{gray}{\textit{Grammar Checkers}}
\vspace{5pt}

\noindent 32. Grammar Checkers adalah alat atau perangkat lunak yang dirancang untuk mendeteksi dan memperbaiki kesalahan ejaan dan tata bahasa dalam teks secara otomatis, sehingga membantu meningkatkan kualitas tulisan.

Apakah Anda pernah menggunakan aplikasi seperti ini? \newline
\textcolor{gray}{\textit{32. Grammar Checkers are tools or software designed to detect and correct spelling and grammar errors in text automatically, thereby helping to improve the quality of writing.
Have you ever used an application like this?}}
\begin{mcq}
    \item Ya \textcolor{blue}{643 (74.7\%)} \newline
    \textcolor{gray}{\textit{Yes}}
    \item Tidak \textcolor{blue}{218 (25.3\%)} \newline
    \textcolor{gray}{\textit{No}}
\end{mcq}
\vspace{15pt}

\noindent 33. Seberapa pentingkah Grammar Checkers bahasa daerah untuk kebutuhan Anda? \newline
\textcolor{gray}{\textit{33. How important is regional language Grammar Checkers for your needs?}}
\begin{mcq}
    \item Sangat penting \textcolor{blue}{329 (38.2\%)} \newline
    \textcolor{gray}{\textit{Very important}}
    \item Penting \textcolor{blue}{316 (36.7\%)} \newline
    \textcolor{gray}{\textit{Important}}
    \item Tidak terlalu penting \textcolor{blue}{173 (20.1\%)} \newline
    \textcolor{gray}{\textit{Not very important}}
    \item Tidak penting \textcolor{blue}{43 (5.0\%)} \newline
    \textcolor{gray}{\textit{Not important}}
\end{mcq}
\vspace{15pt}

\noindent 34. Dimana Anda ingin melihat atau menggunakan Grammar Checkers untuk bahasa daerah Anda? \newline
\textcolor{gray}{\textit{34. Where would you like to see or use Grammar Checkers for your regional language?}}
\begin{multianswer}
    \item Aplikasi ponsel \textcolor{blue}{608 (70.6\%)} \newline
    \textcolor{gray}{\textit{Mobile apps}}
    \item Platform sosial media \textcolor{blue}{288 (33.4\%)} \newline
    \textcolor{gray}{\textit{Social media platforms}}
    \item Situs web \textcolor{blue}{445 (51.7\%)} \newline
    \textcolor{gray}{\textit{Websites}}
    \item Dokumen digital (PDF, word) \textcolor{blue}{237 (27.5\%)} \newline
    \textcolor{gray}{\textit{Digital documents (PDF, word)}}
    \item Platform pembelajaran online \textcolor{blue}{220 (25.6\%)} \newline
    \textcolor{gray}{\textit{Online learning platforms}}
    \item Sistem di tempat kerja \textcolor{blue}{163 (18.9\%)} \newline
    \textcolor{gray}{\textit{Workplace systems}}
    \item Saat bepergian atau di tempat umum \textcolor{blue}{163 (18.9\%)} \newline
    \textcolor{gray}{\textit{While traveling or in public}}
    \item Tidak tertarik \textcolor{blue}{72 (8.4\%)} \newline
    \textcolor{gray}{\textit{Not interested}}
\end{multianswer}
\vspace{15pt}

\noindent \textbf{Mesin Pencarian} \newline
\textcolor{gray}{\textit{Information Retrieval}}
\vspace{5pt}

\noindent 35. Apakah Anda pernah menggunakan teknologi mesin pencarian informasi, seperti Google Search? \newline
\textcolor{gray}{\textit{35. Have you ever used information search engine technology, such as Google Search?}}
\begin{mcq}
    \item Ya \textcolor{blue}{847 (98.4\%)} \newline
    \textcolor{gray}{\textit{Yes}}
    \item Tidak \textcolor{blue}{14 (1.6\%)} \newline
    \textcolor{gray}{\textit{No}}
\end{mcq}
\vspace{15pt}

\noindent 36. Menurut Anda, seberapa pentingkah teknologi mesin pencarian informasi untuk bahasa daerah? \newline
\textcolor{gray}{\textit{36. In your opinion, how important is information search engine technology for regional languages?}}
\begin{mcq}
    \item Sangat penting \textcolor{blue}{556 (64.6\%)} \newline
    \textcolor{gray}{\textit{Very important}}
    \item Penting \textcolor{blue}{250 (29.0\%)} \newline
    \textcolor{gray}{\textit{Important}}
    \item Tidak terlalu penting \textcolor{blue}{49 (5.7\%)} \newline
    \textcolor{gray}{\textit{Not very important}}
    \item Tidak penting \textcolor{blue}{6 (0.7\%)} \newline
    \textcolor{gray}{\textit{Not important}}
\end{mcq}
\vspace{15pt}

\noindent \textbf{Asisten Digital} \newline
\textcolor{gray}{\textit{Digital Assistant}}
\vspace{5pt}

\noindent 37. Asisten digital adalah perangkat lunak berbasis kecerdasan buatan yang membantu pengguna menyelesaikan tugas sehari-hari melalui perintah suara atau teks, seperti menjawab pertanyaan, mengatur jadwal, dan mengontrol perangkat pintar. Contohnya adalah: ChatBot, Siri, Alexa, dan Google Assistant.

Apakah Anda pernah menggunakan aplikasi seperti ini? \newline
\textcolor{gray}{\textit{37. A digital assistant is artificial intelligence-based software that helps users complete everyday tasks through voice or text commands, such as answering questions, setting schedules, and controlling smart devices. Examples are: ChatBot, Siri, Alexa, and Google Assistant.
Have you ever used an application like this?}}
\begin{mcq}
    \item Ya \textcolor{blue}{679 (78.9\%)} \newline
    \textcolor{gray}{\textit{Yes}}
    \item Tidak \textcolor{blue}{182 (21.1\%)} \newline
    \textcolor{gray}{\textit{No}}
\end{mcq}
\vspace{15pt}

\noindent 38. Seberapa pentingkah asisten digital bahasa daerah untuk kebutuhan Anda? \newline
\textcolor{gray}{\textit{38. How important is a regional language digital assistant for your needs?}}
\begin{mcq}
    \item Sangat penting \textcolor{blue}{286 (33.2\%)} \newline
    \textcolor{gray}{\textit{Very important}}
    \item Penting \textcolor{blue}{330 (38.3\%)} \newline
    \textcolor{gray}{\textit{Important}}
    \item Tidak terlalu penting \textcolor{blue}{201 (23.3\%)} \newline
    \textcolor{gray}{\textit{Not very important}}
    \item Tidak penting \textcolor{blue}{44 (5.1\%)} \newline
    \textcolor{gray}{\textit{Not important}}
\end{mcq}
\vspace{15pt}

\noindent 39. Untuk keperluan apa Anda ingin menggunakan asisten digital yang mendukung bahasa daerah Anda? \newline
\textcolor{gray}{\textit{39. For what purposes would you want to use a digital assistant that supports your regional language?}}
\begin{multianswer}
    \item Konsultasi kesehatan \textcolor{blue}{188 (21.8\%)} \newline
    \textcolor{gray}{\textit{Health consultation}}
    \item Curhat masalah pribadi \textcolor{blue}{150 (17.4\%)} \newline
    \textcolor{gray}{\textit{Sharing personal problems}}
    \item Hiburan \textcolor{blue}{316 (36.7\%)} \newline
    \textcolor{gray}{\textit{Entertainment}}
    \item Membantu belajar / pendidikan \textcolor{blue}{514 (59.7\%)} \newline
    \textcolor{gray}{\textit{Help with learning/education}}
    \item Mencari informasi \textcolor{blue}{604 (70.2\%)} \newline
    \textcolor{gray}{\textit{Searching for information}}
    \item Menuliskan teks seperti surat \textcolor{blue}{263 (30.5\%)} \newline
    \textcolor{gray}{\textit{Writing text like a letter}}
    \item Memperbaiki penulisan teks \textcolor{blue}{346 (40.2\%)} \newline
    \textcolor{gray}{\textit{Correcting text writing}}
    \item Tidak perlu \textcolor{blue}{76 (8.8\%)} \newline
    \textcolor{gray}{\textit{Not necessary}}
    \item Lainnya \textcolor{blue}{24 (2.8\%)} \newline
    \textcolor{gray}{\textit{Other}}
\end{multianswer}
\vspace{15pt}

\noindent 40. Asisten digital juga bisa membaca gambar dan video. Apakah menurut Anda penting memiliki Asisten digital berbahasa daerah yang bisa memahami gambar dan video yang berkaitan dengan budaya Anda? \newline
\textcolor{gray}{\textit{40. A digital assistant can also read images and videos. Do you think it is important to have a regional language digital assistant that can understand images and videos related to your culture?}}
\begin{mcq}
    \item Sangat penting \textcolor{blue}{352 (40.9\%)} \newline
    \textcolor{gray}{\textit{Very important}}
    \item Penting \textcolor{blue}{379 (44.0\%)} \newline
    \textcolor{gray}{\textit{Important}}
    \item Tidak terlalu penting \textcolor{blue}{108 (12.5\%)} \newline
    \textcolor{gray}{\textit{Not very important}}
    \item Tidak penting \textcolor{blue}{22 (2.6\%)} \newline
    \textcolor{gray}{\textit{Not important}}
\end{mcq}

\rule{0.2\textwidth}{0.4pt}
\vspace{5pt}

\noindent \textbf{Privasi dan Kredibilitas} \newline
\textcolor{gray}{\textit{Privacy and Credibility}} 
\vspace{5pt}

\noindent 41. Untuk mengembangkan teknologi bahasa daerah, diperlukan banyak data teks dan audio digital dalam bahasa tersebut.
Sebagai contoh, peneliti mungkin akan mengumpulkan dan menganalisis data teks dan audio yang tersedia secara publik di media sosial Anda yang menggunakan bahasa daerah.
Apakah hal ini membuat Anda merasa terganggu? \newline
\textcolor{gray}{\textit{41. To develop regional language technology, a lot of digital text and audio data in that language is needed.
For example, researchers might collect and analyze publicly available text and audio data on your social media that uses your regional language.
Does this bother you?}}
\begin{mcq}
    \item Saya merasa terganggu jika data teks tersebut digunakan untuk pengembangan teknologi bahasa daerah \textcolor{blue}{30 (3.5\%)} \newline
    \textcolor{gray}{\textit{I feel disturbed if the text data is used for the development of regional language technology}}
    \item Saya merasa terganggu jika data audio tersebut digunakan untuk pengembangan teknologi bahasa daerah \textcolor{blue}{29 (3.4\%)} \newline
    \textcolor{gray}{\textit{I feel disturbed if the audio data is used for the development of regional language technology}}
    \item Saya merasa terganggu jika data teks dan audio tersebut digunakan untuk pengembangan teknologi bahasa daerah \textcolor{blue}{36 (4.2\%)} \newline
    \textcolor{gray}{\textit{I feel disturbed if the text and audio data are used for the development of regional language technology}}
    \item Saya tidak merasa terganggu karena data tersebut tersedia secara publik \textcolor{blue}{766 (89.0\%)} \newline
    \textcolor{gray}{\textit{I do not feel disturbed because the data is publicly available}}
\end{mcq}
\vspace{15pt}

\noindent 42. Apakah Anda merasa teknologi kecerdasan buatan yang sudah ada memberikan perlindungan terhadap data pribadi Anda secara memadai? \newline
\textcolor{gray}{\textit{42. Do you feel that existing artificial intelligence technologies provide adequate protection for your personal data?}}
\begin{mcq}
    \item Ya \textcolor{blue}{214 (24.9\%)} \newline
    \textcolor{gray}{\textit{Yes}}
    \item Tidak \textcolor{blue}{379 (44.0\%)} \newline
    \textcolor{gray}{\textit{No}}
    \item Tidak tahu \textcolor{blue}{268 (31.1\%)} \newline
    \textcolor{gray}{\textit{Do not know}}
\end{mcq}
\vspace{15pt}

\noindent 43. Saat menggunakan teknologi bahasa seperti Google Search, Siri, dan Google Assistant, apakah Anda sudah pernah mendengar tentang isu privasi dan keamanan? Misalnya, tidak menyebutkan atau menuliskan data pribadi ke asisten digital seperti ChatGPT? \newline
\textcolor{gray}{\textit{43. When using language technologies such as Google Search, Siri, and Google Assistant, have you heard about privacy and security issues? For example, not mentioning or writing personal data to digital assistants such as ChatGPT?}}
\begin{mcq}
    \item Sangat tahu \textcolor{blue}{140 (16.3\%)} \newline
    \textcolor{gray}{\textit{Very aware}}
    \item Cukup tahu \textcolor{blue}{354 (41.1\%)} \newline
    \textcolor{gray}{\textit{Aware}}
    \item Tidak terlalu tahu \textcolor{blue}{216 (25.1\%)} \newline
    \textcolor{gray}{\textit{Not too aware}}
    \item Tidak tahu \textcolor{blue}{151 (17.5\%)} \newline
    \textcolor{gray}{\textit{Not aware}}
\end{mcq}
\vspace{15pt}

\noindent 44. Apakah Anda pernah menanyakan masalah kesehatan kepada asisten digital seperti ChatGPT? \newline
\textcolor{gray}{\textit{44. Have you ever asked a digital assistant such as ChatGPT about health problems?}}
\begin{mcq}
    \item Pernah \textcolor{blue}{278 (32.3\%)} \newline
    \textcolor{gray}{\textit{I have}}
    \item Tidak pernah \textcolor{blue}{583 (67.7\%)} \newline
    \textcolor{gray}{\textit{I have not}}
\end{mcq}
\vspace{15pt}

\noindent 45. Seberapa sering Anda melakukan verifikasi kebeneran informasi yang diberikan oleh teknologi bahasa seperti ChatGPT? \newline
\textcolor{gray}{\textit{45. How often do you verify the accuracy of information provided by language technology such as ChatGPT?}}
\begin{mcq}
    \item Selalu \textcolor{blue}{130 (15.1\%)} \newline
    \textcolor{gray}{\textit{Always}}
    \item Sering \textcolor{blue}{262 (30.4\%)} \newline
    \textcolor{gray}{\textit{Often}}
    \item Jarang \textcolor{blue}{274 (31.8\%)} \newline
    \textcolor{gray}{\textit{Seldom}}
    \item Tidak pernah \textcolor{blue}{195 (22.6\%)} \newline
    \textcolor{gray}{\textit{Never}}
\end{mcq}
\vspace{15pt}

\noindent 46. Apakah Anda tahu bahwa informasi yang diberikan oleh asisten digital seperti ChatGPT tidak selalu benar dan bisa sepenuhnya salah? \newline
\textcolor{gray}{\textit{46. Do you know that information provided by digital assistants such as ChatGPT is not always correct and can be completely wrong?}}
\begin{mcq}
    \item Sangat tahu \textcolor{blue}{311 (36.1\%)} \newline
    \textcolor{gray}{\textit{Very aware}}
    \item Cukup tahu \textcolor{blue}{323 (37.5\%)} \newline
    \textcolor{gray}{\textit{Aware}}
    \item Tidak terlalu tahu \textcolor{blue}{109 (12.7\%)} \newline
    \textcolor{gray}{\textit{Not too aware}}
    \item Tidak tahu \textcolor{blue}{118 (13.7\%)} \newline
    \textcolor{gray}{\textit{Not aware}}
\end{mcq}
\vspace{15pt}

\noindent 47. \textcolor{red}{Pilihlah opsi jawaban Stroberi} \newline
\textcolor{gray}{\textit{47. Choose the Strawberry answer option}}
\begin{mcq}
    \item Apel* \textcolor{blue}{10 (1.2\%)} \newline
    \textcolor{gray}{\textit{Apple}}
    \item Pisang* \textcolor{blue}{4 (0.5\%)} \newline
    \textcolor{gray}{\textit{Banana}}
    \item Jeruk* \textcolor{blue}{4 (0.5\%)} \newline
    \textcolor{gray}{\textit{Orange}}
    \item Stroberi \textcolor{blue}{832 (96.6\%)} \newline
    \textcolor{gray}{\textit{Strawberry}}
    \item Semangka* \textcolor{blue}{11 (1.3\%)} \newline
    \textcolor{gray}{\textit{Watermelon}}
\end{mcq}
\textit{note: \textcolor{red}{*we omit the responses from analysis}}
\vspace{15pt}

\noindent 48. Saat menggunakan teknologi bahasa, apakah Anda sudah pernah mendengar tentang isu bias? Misalnya:

(1) Bias terhadap gender: komputer mengasumsikan bahwa dokter adalah laki-laki dan perawat adalah perempuan. Padahal terdapat dokter perempuan dan perawat laki-laki.
(2) Bias terhadap agama/politik: komputer mencerminkan prasangka terhadap agama/politik tertentu sehingga menyudutkan kalangan tertentu. \newline
\textcolor{gray}{\textit{48. When using language technology, have you ever heard of bias issues? For example:
(1) Gender bias: computers assume that doctors are male and nurses are female. In fact, there are female doctors and male nurses.
(2) Bias against religion/politics: computers reflect prejudice against certain religions/politics, thus cornering certain groups.}}
\begin{mcq}
    \item Sangat tahu \textcolor{blue}{138 (16.0\%)} \newline
    \textcolor{gray}{\textit{Very aware}}
    \item Cukup tahu \textcolor{blue}{335 (38.9\%)} \newline
    \textcolor{gray}{\textit{Aware}}
    \item Tidak terlalu tahu \textcolor{blue}{216 (25.1\%)} \newline
    \textcolor{gray}{\textit{Not too aware}}
    \item Tidak tahu \textcolor{blue}{172 (20.0\%)} \newline
    \textcolor{gray}{\textit{Not aware}}
\end{mcq}
\vspace{15pt}

\noindent 49. Tulis isu lain yang ingin Anda sampaikan terkait teknologi bahasa seperti ChatBot, asisten digital, mesin penerjemah dll. \newline
\textcolor{gray}{\textit{49. Write other issues that you want to convey regarding language technology such as ChatBot, digital assistants, machine translators, etc.}} \newline
\textcolor{blue}{861 write-in answers}

\rule{0.2\textwidth}{0.4pt}
\vspace{5pt}

\noindent \textbf{Privasi dan Kredibilitas} \newline
\textcolor{gray}{\textit{Privacy and Credibility}} 
\vspace{5pt}

\noindent 50. Secara umum, bagaimana antusiasme Anda terhadap pengembangan teknologi bahasa untuk bahasa daerah Anda? Apakah Anda memiliki kekhawatiran atau ketidaksukaan terkait pengembangannya? \newline
\textcolor{gray}{\textit{50. In general, how enthusiastic are you about the development of language technology for your regional language? Do you have any concerns or dislikes regarding its development?}}
\begin{mcq}
    \item Saya antusias dan tidak khawatir \textcolor{blue}{512 (59.5\%)} \newline
    \textcolor{gray}{\textit{I am enthusiastic and not worried}}
    \item Saya antusias dan sedikit khawatir \textcolor{blue}{287 (33.3\%)} \newline
    \textcolor{gray}{\textit{I am enthusiastic and a little worried}}
    \item Saya tidak antusias, namun sedikit khawatir \textcolor{blue}{26 (3.0\%)} \newline
    \textcolor{gray}{\textit{I am not enthusiastic, but a little worried}}
    \item Saya tidak antusias dan tidak khawatir \textcolor{blue}{36 (4.2\%)} \newline
    \textcolor{gray}{\textit{I am neither enthusiastic nor worried}}
\end{mcq}
\vspace{15pt}


\section{Details of Variations of Importance Scores}
\label{sec:variation-importance-score}
Table~\ref{tab:survey-data} presents the importance scores across various categories, along with their standard deviations for statistical analysis. We calculate the machine translation (MT) preferences using map in Table~\ref{tab:mt-mapping}, making it uniform with the other LTs. It is important to note that the standard deviations are influenced by the nature of the response options, which were limited to four choices: Very Important ($3/3$), Important ($2/3$), Not Too Important ($1/3$), and Not Important ($0/3$). This scale means that each option differs by increments of 0.33. As shown in Table~\ref{tab:survey-data}, the results are generally consistent, except for Moribund languages, which have a standard deviation greater than 0.33, likely due to the smaller number of participants in that category.

\begin{table*}[h]
    \small
    \centering
    \begin{tabular}{|l|c|c|c|c|c|c|c|}
        \hline
        \textbf{Categories} & \textbf{MT} & \textbf{STT} & \textbf{TTS} & \textbf{GC} & \textbf{IR} & \textbf{DA} \\
        \hline
        full & 0.771 \textsubscript{($\pm0.25$)} & 0.678 \textsubscript{($\pm0.28$)} & 0.684 \textsubscript{($\pm0.28$)} & 0.696 \textsubscript{($\pm0.29$)} & 0.860 \textsubscript{($\pm0.21$)} & 0.664 \textsubscript{($\pm0.29$)} \\ \hline
        aware of bias & 0.766 \textsubscript{($\pm0.27$)} & 0.692 \textsubscript{($\pm0.28$)}  & 0.699 \textsubscript{($\pm0.28$)}  & 0.709 \textsubscript{($\pm0.29$)}  & 0.868 \textsubscript{($\pm0.21$)}  & 0.678 \textsubscript{($\pm0.29$)}  \\
        not aware of bias & 0.777 \textsubscript{($\pm0.24$)}  & 0.661 \textsubscript{($\pm0.28$)}  & 0.664 \textsubscript{($\pm0.28$)} & 0.681 \textsubscript{($\pm0.3$)}  & 0.851 \textsubscript{($\pm0.21$)}  & 0.646 \textsubscript{($\pm0.29$)}  \\ \hline
        aware of privacy & 0.759 \textsubscript{($\pm0.27$)}  & 0.673 \textsubscript{($\pm0.29$)}  & 0.682 \textsubscript{($\pm0.29$)}  & 0.695 \textsubscript{($\pm0.3$)}  & 0.864 \textsubscript{($\pm0.21$)}  & 0.662 \textsubscript{($\pm0.3$)}  \\
        not aware of privacy & 0.786 \textsubscript{($\pm0.23$)}  & 0.685 \textsubscript{($\pm0.26$)}  & 0.685 \textsubscript{($\pm0.26$)}  & 0.700 \textsubscript{($\pm0.28$)}  & 0.855 \textsubscript{($\pm0.21$)}  & 0.667 \textsubscript{($\pm0.28$)}  \\ \hline
        geo: west Indonesia & 0.762 \textsubscript{($\pm0.26$)}  & 0.675 \textsubscript{($\pm0.28$)}  & 0.661 \textsubscript{($\pm0.28$)}  & 0.675 \textsubscript{($\pm0.30$)} & 0.848 \textsubscript{($\pm0.22$)}  & 0.634 \textsubscript{($\pm0.29$)}  \\
        geo: east Indonesia & 0.792 \textsubscript{($\pm0.23$)}  & 0.729 \textsubscript{($\pm0.27$)}  & 0.737 \textsubscript{($\pm0.26$)}  & 0.748 \textsubscript{($\pm0.28$)}  & 0.889 \textsubscript{($\pm0.19$)}  & 0.737 \textsubscript{($\pm0.28$)}  \\ \hline
        edu: high school & 0.721 \textsubscript{($\pm0.28$)}  & 0.664 \textsubscript{($\pm0.28$)}  & 0.677 \textsubscript{($\pm0.29$)}  & 0.694 \textsubscript{($\pm0.29$)}  & 0.878 \textsubscript{($\pm0.18$)}  & 0.679 \textsubscript{($\pm0.29$)}  \\
        edu: undergraduate & 0.792 \textsubscript{($\pm0.22$)}  & 0.688 \textsubscript{($\pm0.27$)}  & 0.687 \textsubscript{($\pm0.27$)}  & 0.700 \textsubscript{($\pm0.29$)}  & 0.868 \textsubscript{($\pm0.21$)}  & 0.674 \textsubscript{($\pm0.29$)}  \\
        edu: graduate & 0.765 \textsubscript{($\pm0.28$)}  & 0.671 \textsubscript{($\pm0.29$)}  & 0.682 \textsubscript{($\pm0.29$)}  & 0.693 \textsubscript{($\pm0.30$)}  & 0.841 \textsubscript{($\pm0.22$)}  & 0.644 \textsubscript{($\pm0.29$)}  \\ \hline
        lang: stable & 0.763 \textsubscript{($\pm0.26$)}  & 0.663 \textsubscript{($\pm0.28$)}  & 0.668 \textsubscript{($\pm0.28$)}  & 0.684 \textsubscript{($\pm0.29$)}  & 0.843 \textsubscript{($\pm0.22$)}  & 0.642 \textsubscript{($\pm0.29$)}  \\
        lang: endangered & 0.804 \textsubscript{($\pm0.22$)}  & 0.731 \textsubscript{($\pm0.27$)}  & 0.723 \textsubscript{($\pm0.28$)}  & 0.740 \textsubscript{($\pm0.28$)}  & 0.896 \textsubscript{($\pm0.19$)}  & 0.723 \textsubscript{($\pm0.29$)}  \\
        lang: moribund & 0.608 \textsubscript{($\pm0.31$)}  & 0.490 \textsubscript{($\pm0.33$)}  & 0.510 \textsubscript{($\pm0.33$)}  & 0.451 \textsubscript{($\pm0.34$)}  & 0.863 \textsubscript{($\pm0.20$)}  & 0.569 \textsubscript{($\pm0.34$)}  \\ \hline
        familiar to LT & 0.775 \textsubscript{($\pm0.26$)}  & 0.714 \textsubscript{($\pm0.26$)}  & 0.733 \textsubscript{($\pm0.25$)}  & 0.724 \textsubscript{($\pm0.29$)}  & 0.864 \textsubscript{($\pm0.21$)}  & 0.705 \textsubscript{($\pm0.28$)}  \\
        $\sim$familiar to LT & 0.713 \textsubscript{($\pm0.22$)}  & 0.560 \textsubscript{($\pm0.30$)}  & 0.551 \textsubscript{($\pm0.30$)}  & 0.606 \textsubscript{($\pm0.28$)}  & 0.576 \textsubscript{($\pm0.29$)}  & 0.509 \textsubscript{($\pm0.30$)}  \\ \hline
        gen z & 0.763 \textsubscript{($\pm0.26$)}  & 0.669 \textsubscript{($\pm0.28$)}  & 0.685 \textsubscript{($\pm0.27$)}  & 0.708 \textsubscript{($\pm0.30$)}  & 0.878 \textsubscript{($\pm0.19$)}  & 0.685 \textsubscript{($\pm0.29$)}  \\
        gen millennial & 0.773 \textsubscript{($\pm0.26$)}  & 0.689 \textsubscript{($\pm0.28$)}  & 0.685 \textsubscript{($\pm0.28$)}  & 0.685 \textsubscript{($\pm0.30$)}  & 0.855 \textsubscript{($\pm0.21$)}  & 0.658 \textsubscript{($\pm0.29$)}  \\
        gen x boomer & 0.782 \textsubscript{($\pm0.20$)}  & 0.658 \textsubscript{($\pm0.27$)}  & 0.671 \textsubscript{($\pm0.28$)}  & 0.722 \textsubscript{($\pm0.25$)}  & 0.829 \textsubscript{($\pm0.25$)}  & 0.641 \textsubscript{($\pm0.29$)}  \\
        \hline
    \end{tabular}
    \caption{Importance scores along with the standard deviations across demographic and awareness categories.}
    \label{tab:survey-data}
\end{table*}

\begin{table}
    \centering
    \begin{tabular}{l|l} \hline 
         \textbf{Category}& \textbf{Criteria}\\ \hline 
         Very Important & Select 3+ options \\  
         Important & Selects 1-2 option(s) \\
         Not Important & Selects 0 options \\ \hline
    \end{tabular}
    \caption{The mapping of user preferences towards MT. We are mapping the user's answer this way to have a uniform category with the other LTs, while having more insight into what the user exactly wants in MT.}
    \label{tab:mt-mapping}
\end{table}

\section{The Division of West and East Indonesia based on Wikipedia}
\label{sec:indo_barat_timur}
We aggregated the results based on several criteria, including clustering Indonesia into West and East regions. We referred to relevant Wikipedia pages\footnote{\url{https://id.wikipedia.org/wiki/Indonesia_Barat}, \url{https://id.wikipedia.org/wiki/Indonesia_Timur}} for a straightforward classification of provinces, as well as classified based on their historical contexts and economic disparities. Table \ref{tab:west-east-indo} presents the distribution between West and East Indonesia, followed by the respondent count for each province.


\begin{table}
    \centering
    \begin{tabular}{|l|l|} \hline 
         \textbf{West Indonesia}& \textbf{East Indonesia}\\ \hline 
         East Java (112)&South Sulawesi (67) \\  
         West Java (111)&NTB (37) \\
        Central Java (72)&NTT (34) \\ 
        West Sumatera (54)&Bali (32) \\
        Aceh (37)&Central Sulawesi (32)\\
        North Sumatera (33)&S.E. Sulawesi (14) \\
        DI Yogyakarta (29)&Papua (8) \\
        Jakarta (29)&North Sulawesi (3) \\
        Riau (18)&West Sulawesi (3) \\
        Jambi (17)&Highland Papua (3) \\ 
        West Kalimantan (13)&Gorontalo (1) \\
        South Sumatera (12)&West Papua (1)\\
        Lampung (6)&Central Papua (1) \\
        Bengkulu (6)&Maluku (1) \\
        South Kalimantan (6)&S.W. Papua (0) \\
        Banten (5)&South Papua (0) \\
        East Kalimantan (4)&North Maluku (0) \\ 
        Ctrl. Kalimantan (4) &\\ 
        Riau Islands (3) &\\ 
        Bangka Belitung (2) &\\ 
        North Kalimantan (1) &\\ \hline
        \textbf{Total}=574 &\textbf{Total}=237 \\ \hline
    \end{tabular}
    \caption{The division and the valid respondent count based on province location (West \& East Indonesia).}
    \label{tab:west-east-indo}
\end{table}

\section{Language Level Aggregation}
\label{sec:lang_level_aggregation}
\citet{ethnologue2024} established a language taxonomy based on real-world usage. The taxonomy consists of nine language status levels, ranging from International to Extinct language \footnote{\url{https://www.ethnologue.com/methodology/\#language-status}}:

\begin{itemize}
    \item 0. International: The language is widely used between nations in trade, knowledge exchange, and international policy. \textit{Not applicable in our survey}
    \item 1.	National: The language is used in education, work, mass media, and government at the national level. \textit{Not applicable in our survey}
    \item 2. Provincial:	The language is used in education, work, mass media, and government within major administrative subdivisions of a nation. \textit{Not applicable in our survey}
    \item 3. Wider Communication: The language is used in work and mass media without official status to transcend language differences across a region.
    \item 4. Educational: The language is in vigorous use, with standardization and literature being sustained through a widespread system of institutionally supported education.
    \item 5. Developing: The language is in vigorous use, with literature in a standardized form being used by some though this is not yet widespread or sustainable.
    \item 6a. Vigorous: The language is used for face-to-face communication by all generations and the situation is sustainable.
    \item 6b. Threatened: The language is used for face-to-face communication within all generations, but it is losing users.
    \item 7. Shifting: The child-bearing generation can use the language among themselves, but it is not being transmitted to children.
    \item 8a. Moribund: The only remaining active users of the language are members of the grandparent generation and older.
    \item 8b. Nearly Extinct: The only remaining users of the language are members of the grandparent generation or older who have little opportunity to use the language.
    \item 9. Dormant: The language serves as a reminder of heritage identity for an ethnic community, but no one has more than symbolic proficiency.
    \item 10. Extinct: The language is no longer used, and no one retains a sense of ethnic identity associated with the language. \textit{Not applicable in our survey}
\end{itemize}

However, for ease of analysis, we consolidated these 13 levels into 3 broader categories. Table \ref{tab:lang-level-aggregation} presents our classification along with the languages covered in the survey.

\begin{table*}
    \centering
    \renewcommand{\arraystretch}{1}
    \begin{tabular}{|p{4cm}|p{300pt}|} \hline 
         \textbf{Language Level} & \textbf{Covered Languages} \\ \hline 
         \textbf{Stable} Language (\textit{Ethnologue language level 3-5}) & 
         Javanese (245), Sunda (105), Bugis (64), Minangkabau (62), Bali (30), Kaili Ledo (13), Musi (9), Madura (7), Banjar (6), Toraja-sadan (6), Lamaholot (4), Malay-manado (3), Ngaju (3), Chinese-mandarin (3), Mandar (2), Kendayan (1), Moma (1), Nias (1), Malay-kupang (1) \\ \hline 
         
         \textbf{Threatened} Language (\textit{Ethnologue language level 6a-6b}) & 
         Aceh (33), Sasak (22), Malay (20), Malay-jambi (13), Batak simalungun (12), Batak toba (7), Hawu (7), Saluan (6), Bima (5), Lampung nyo (4), Sumbawa (4), Tolaki (4), Malay-central (4), Tetun (4), Uab meto (3), Manggarai (3), Biak (3), Muna (3), Kambera (3), Tukang besi south (2), Li’o (2), Batak karo (2), Moronene (2), Pamona (2), Konjo-coastal (2), Osing (2), Padoe (1), Bahau (1), Sika (1), 
         Betawi (1), Batak mandailing (1), Ende (1), Batak alas-kluet (1), Gayo (1), Bangka (1), Malay-tenggarong kutai (1), Bakati’ (1), Tii (1), Gorontalo (1), Sentani (1), Nalca (1), Ekari (1), Ketengban (1), Ansus (1), Diuwe (1), Rejang (1), Mamuju (1), Cia-cia (1) \\ \hline 
         
         \textbf{Moribund} Language (\textit{Ethnologue language level 7-9}) & 
         Hakka (12), Banggai (3), Andio (2) \\ \hline 
    \end{tabular}
    \caption{Language level classification and the valid respondent count based on each language.}
    \label{tab:lang-level-aggregation}
\end{table*}

\section{Dialect-Based User Preferences}
\label{sec:dialect-appendix}
\begin{figure*}
    \centering
    \includegraphics[width=1\linewidth]{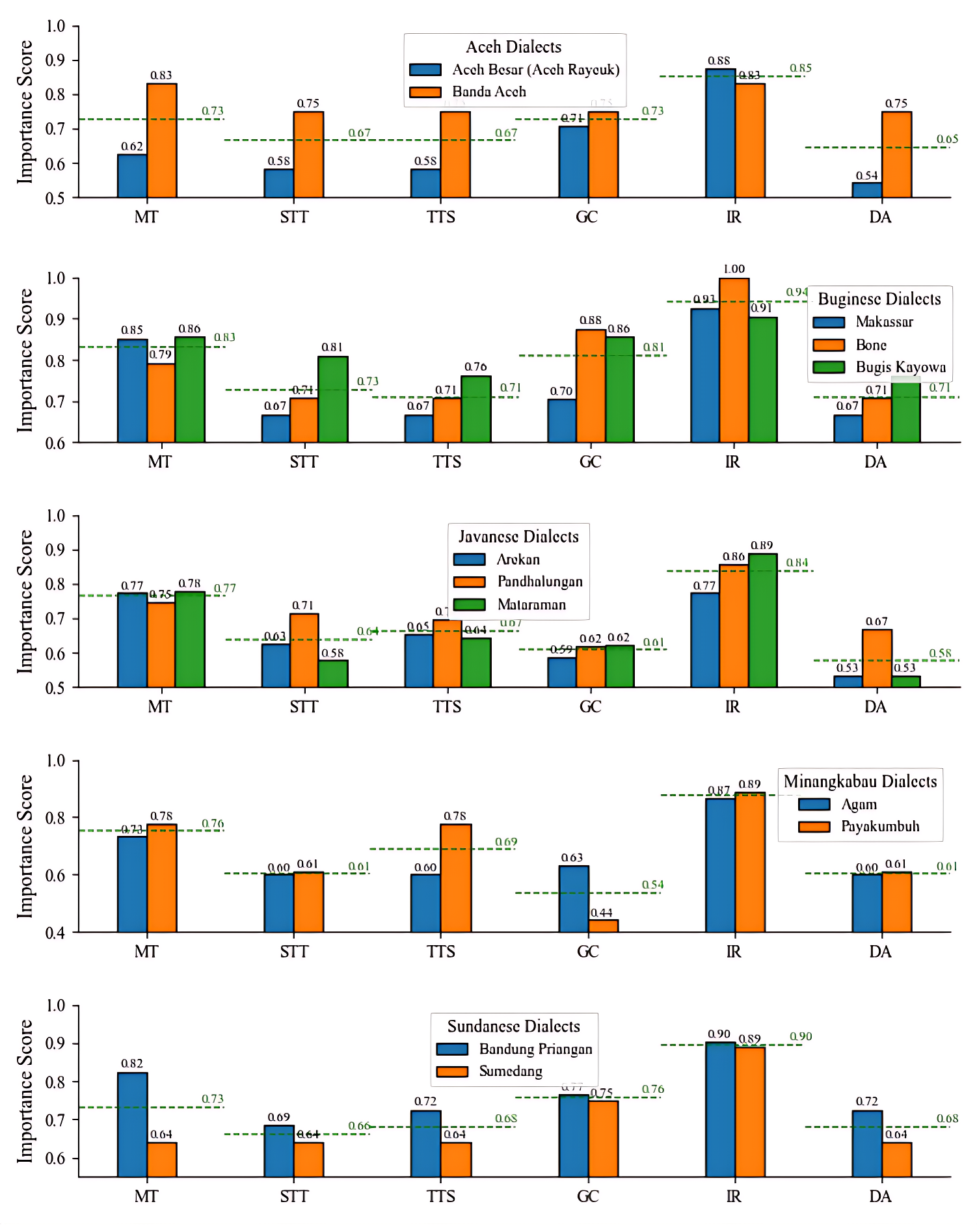}
    \caption{Differences in LT preferences across Aceh, Buginese, Javanese, Minangkabau, and Sundanese dialects (from top to bottom).}
    \label{fig:dialects-full}
\end{figure*}

As discussed in Section \ref{sec:dialects}, dialects also influence how speakers of the same language perceive the need for language technologies (LTs). Due to limited respondent counts, we focused on five languages and their respective dialects: Aceh (Aceh Besar and Banda Aceh dialects), Buginese (Makassar, Bone, and Bugis Kayowa dialects), Javanese (Arekan, Pandhalungan, and Mataraman dialects), Minangkabau (Agam and Payakumbuh dialects), and Sundanese (Bandung Priangan and Sumedang dialects) as shown in Figure \ref{fig:dialects-full}.

Overall, the Banda Aceh, Payakumbuh, and Bandung Priangan dialects stand out as perceiving LTs as more important compared to other dialects within their respective languages. Notably, the Bone dialect in Buginese shows a distinct preference, with speakers prioritizing GC and IR more but showing less interest in MT. In contrast, the Makassar dialect perceives LTs as less important than other Buginese dialects.

However, the reasons behind these trends remain unclear. To fully understand why certain dialects exhibit unique patterns in perceiving LTs, direct dialogue with speakers of each dialect is essential.

\section{How Awareness of Privacy Affects Use Rate}
\label{sec:privacy-use-rate}
\begin{figure}
    \centering
    \includegraphics[width=1\linewidth]{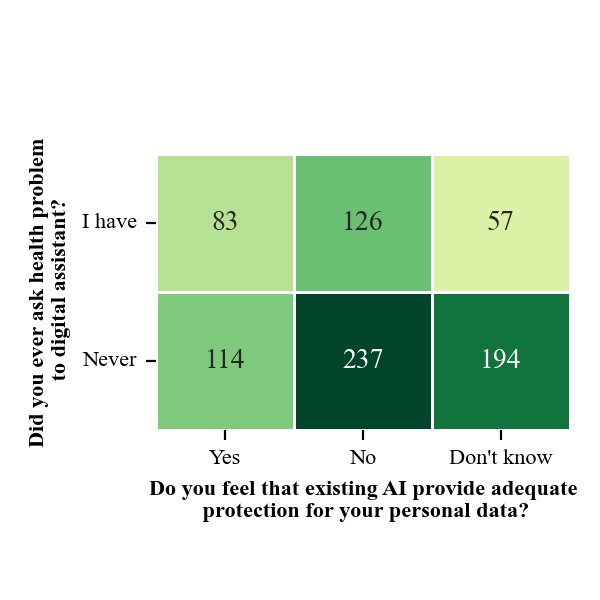}
    \caption{How awareness of privacy affects use rate.}
    \label{fig:awareness-priv-affects-use}
\end{figure}

Figure \ref{fig:awareness-priv-affects-use} illustrates the relationship between respondents' awareness of privacy concerns and their usage rates of language technologies (LTs). Overall, individuals who believe that LTs fail to provide sufficient protection for personal data are less likely to use digital assistants for health-related inquiries, as such information is considered highly sensitive. Similarly, those who remain uncertain about the level of data protection offered by LTs tend to avoid using these technologies for health-related questions altogether.

\section{Familiarity with LTs: Categorized on Generation, Language Level, and Geography}
\label{sec:familiarity-familiar}
\begin{figure*}
    \centering
    \includegraphics[width=0.98\linewidth]{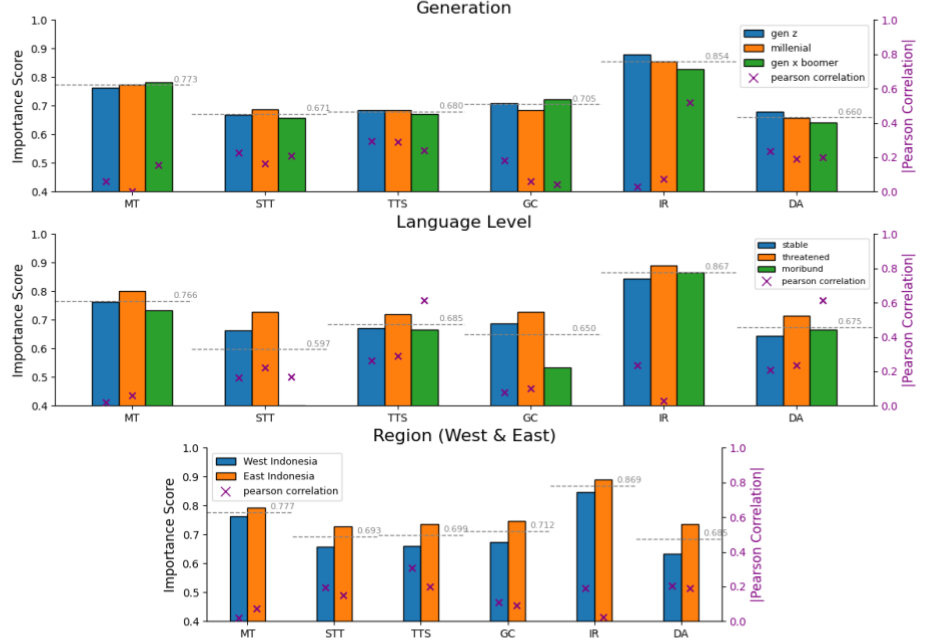}
    \caption{Importance scores of the respondents that are familiar with the LT across several categories: Generation, language level, and region (West \& East Indonesia.), alongside their Pearson correlation.}
    \label{fig:importance-score-gen-langlvl-westeast}
\end{figure*}

Figure \ref{fig:familiar-generation-familiar} illustrates respondents' familiarity with LTs analyzed in this survey, categorized by different factors. Among generations, Gen Z appears to be the most familiar with LTs, while Gen X \& Boomers show the lowest familiarity, likely due to the rapid pace of globalization affecting younger generations more. Additionally, speakers of stable languages tend to have higher LT familiarity compared to others. Geographically, respondents from West Indonesia are more familiar with LTs than those from East Indonesia, likely due to Indonesia's development being concentrated in more populous islands such as Java and Sumatra. In addition, Figure \ref{fig:importance-score-gen-langlvl-westeast} shows the importance scores of the respondents who are familiar with the LT across several categories, followed by the Pearson correlation between the familiarity of LT to its importance score.

\section{Important Score vs Available Resource on Wikipedia}
\label{sec:important-vs-resourcewiki}
\begin{figure}
    \centering
    \includegraphics[width=0.98\linewidth]{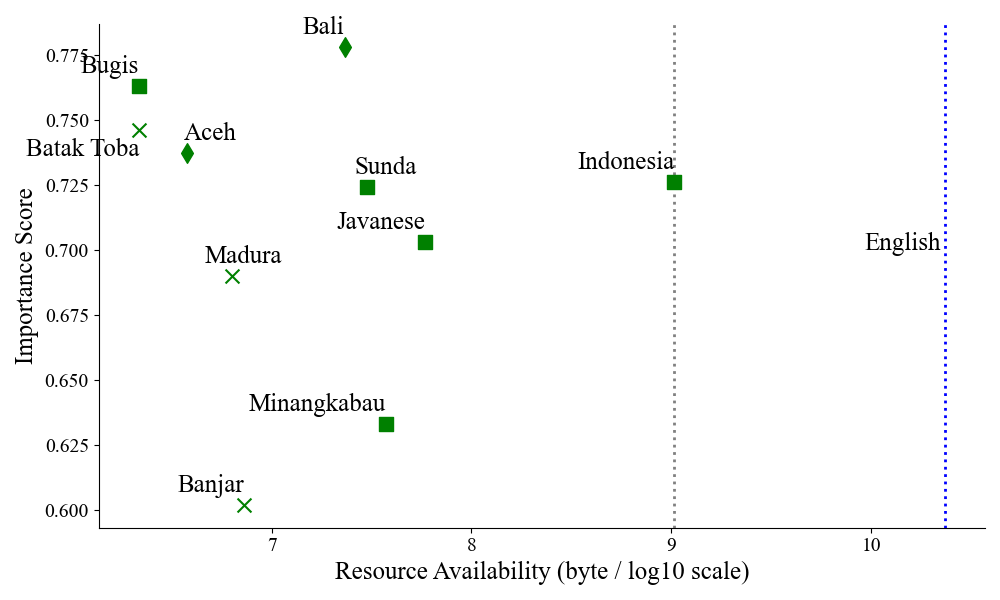}
    \caption{Importance scores and available resources for each supported local Indonesian language on Wikipedia. $\blacksquare$ represents languages that has more than 50 respondents, $\blacklozenge$ 30-50 respondents, and $\times$ is less than 30 respondents.}
    \label{fig:importance-vs-resourcewiki}
\end{figure}

We use Wikipedia data as a common text source for dataset collection. Figure~\ref{fig:importance-vs-resourcewiki} illustrates that despite the high importance scores of several Indonesian local languages, the available resources remain insufficient. Only a few languages—such as Javanese, Sundanese, Balinese, and Minangkabau—have datasets exceeding 10MB (which is still considered tiny). Meanwhile, resources for all other languages remain limited or entirely unsupported.

\section{Current State of Language Technologies for Indonesian Local Languages}
\label{sec:lt-curr-state}
Table \ref{tab:existing-lt-covered-by-google} presents the current state of LTs for Indonesian local languages, using Google as a benchmark. While some languages, such as Javanese and Sundanese, are supported in certain LTs, many other underrepresented languages still lack coverage. Additionally, technologies like TTS and DA have yet to support any Indonesian regional languages. This provides an overview of the development gaps in LTs for these languages.

\begin{table*}
    \centering
    \begin{tabular}{|c|c|c|} \hline 
         \thead{LT}&  \thead{Importance score}& \thead{Local Indonesian Language(s) supported by Google}\\ \hline 
         MT& 0.771 & \makecell{Javanese (jav), Sundanese (sun), Minangkabau (min),\\ Acehnese (ace), Balinese (ban), Batak Karo (btx), Batak Simalungun (bts), \\Batak Toba (bbc), Betawi (bew), Makassar Malay (mfp)}\\ \hline 
         STT& 0.678 & Javanese (jav), Sundanese (sun)\\ \hline
         TTS& 0.684 & \textit{not supported (only available in Indonesian (id))}\\ \hline 
         GC& 0.696 & \makecell{Ambonese Malay (abs), Batak Simalungun (bts), \\ Buginese (bug), Duri (mvp), Hawu (hvn), Makassar Malay (mfp), \\ Toraja-sa'dan (sda), Acehnese (ace), Batak Alas-kluet (btz), \\ Balinese (ban)*, Banjar (bjn), Batak Mandailing (btm), \\Batak Toba (bbc), Betawi (bew), Gorontalo (gor), Jambi Malay (jax), \\Javanese (jav)*, Kutai Malay (vkt), Ledo Kaili (lew), \\Manado Malay (xmm), Mandar (mdr), Minangkabau (min), \\Mongondow (mog), Papuan Malay (pmy), Sasak (sas), Sundanese (sun)}\\ \hline 
         IR& 0.860 & Javanese (jav)**\\ \hline 
         DA& 0.664 & \textit{not supported}***\\ \hline
    \end{tabular}
    \caption{Importance score for each LT and its availability in local Indonesian languages supported by Google. The \textit{italic} importance score only considers the `very important' option. *their script alphabets are also supported **only able to extract entities from document ***Google Assistant (Android handphone \& TV) \& Gemini only available in Indonesian (ind) language.}
    \label{tab:existing-lt-covered-by-google}
\end{table*}

\end{document}